\documentclass[runningheads]{llncs}

\usepackage{eccv}

\usepackage{eccvabbrv}

\usepackage{graphicx}
\usepackage{booktabs}
\usepackage{colortbl}
\usepackage{multirow}

\usepackage[accsupp]{axessibility}  % Improves PDF readability for those with disabilities.

\usepackage{hyperref}

\usepackage{orcidlink}

\makeatletter
\newcommand\figcaption{\def\@captype{figure}\caption}  % for captions of table in minipage
\newcommand\tabcaption{\def\@captype{table}\caption}
\makeatother

\begin{document}

% ---------------------------------------------------------------
% TODO REVIEW: Replace with your title
\title{LEGO: \underline{L}earning \underline{EGO}centric Action Frame Generation via Visual Instruction Tuning}

% TODO REVIEW: If the paper title is too long for the running head, you can set
% an abbreviated paper title here. If not, comment out.
\titlerunning{Learning Egocentric Action Frame Generation}

% TODO FINAL: Replace with your author list. 
% Include the authors' OCRID for the camera-ready version, if at all possible.
\author{Bolin Lai\inst{1,2,\dagger} \and Xiaoliang Dai\inst{1} \and Lawrence Chen\inst{1} \and Guan Pang\inst{1} \\ \and James M. Rehg\inst{3} \and Miao Liu\inst{1}}

% TODO FINAL: Replace with an abbreviated list of authors.
\authorrunning{B. Lai et al.}
% First names are abbreviated in the running head.
% If there are more than two authors, 'et al.' is used.

% TODO FINAL: Replace with your institution list.
\institute{GenAI, Meta \and Georgia Institute of Technology \and
University of Illinois Urbana-Champaign \\ 
{\tt bolin.lai@gatech.edu \ \{xiaoliangdai,lawrencechen,gpang,miaoliu\}@meta.com \ jrehg@illinois.edu}}

\maketitle

\vspace{-0.2cm}
\begin{center}
  \includegraphics[width=\linewidth]{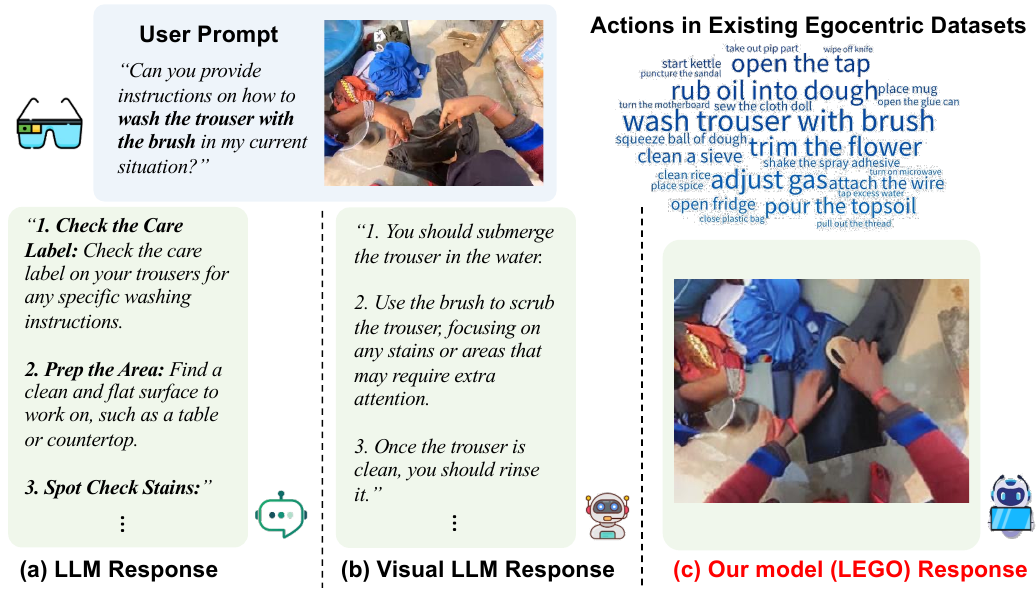}
  \captionof{figure}{When a user performing a complex task asks a large language model (LLM) for instructions (a) on how to complete the steps, she receives a generic answer and has to translate the guidance into her specific situation. If she is wearing a camera, then the prompt can be augmented with an egocentric view of the scene and passed to a Visual LLM (b), and the description is now contextualized to her situation. But she still faces the challenge of parsing a written description. When she uses our novel LEGO model (c), however, the combined image and prompt are used to \emph{automatically generate an image} that provides visual guidance exactly in her situation from the egocentric viewpoint. Now she can complete her task seamlessly!}
  \label{fig:teaser}
\end{center}
\vspace{-0.2cm}

\let\thefootnote\relax\footnotetext{$^\dagger$This work was done during Bolin's internship at GenAI, Meta.}

\begin{abstract}
Generating instructional images of human daily actions from an egocentric viewpoint serves as a key step towards efficient skill transfer. In this paper, we introduce a novel problem -- egocentric action frame generation. The goal is to synthesize an image depicting an action in the user's context (\ie, action frame) by conditioning on a user prompt and an input egocentric image. Notably, existing egocentric action datasets lack the detailed annotations that describe the execution of actions. Additionally, existing diffusion-based image manipulation models are sub-optimal in controlling the state transition of an action in egocentric image pixel space because of the domain gap. To this end, we propose to Learn EGOcentric (LEGO) action frame  generation via visual instruction tuning. First, we introduce a prompt enhancement scheme to generate enriched action descriptions from a visual large language model (VLLM) by visual instruction tuning. Then we propose a novel method to leverage image and text embeddings from the VLLM as additional conditioning to improve the performance of a diffusion model. We validate our model on two egocentric datasets -- Ego4D and Epic-Kitchens. Our experiments show substantial improvement over prior image manipulation models in both quantitative and qualitative evaluation. We also conduct detailed ablation studies and analysis to provide insights in our method. More details of the dataset and code are available on the website (\url{https://bolinlai.github.io/Lego_EgoActGen/}).

\vspace{-0.1cm}
\end{abstract}

% -------------------------- Introduction -------------------------------
\section{Introduction}
\label{sec:intro}

The emergence of Large Language Models (LLMs)~\cite{chowdhery2022palm,brown2020language,raffel2020exploring,zhang2022opt}, such as ChatGPT, has revolutionized the transfer of knowledge. However, an LLM alone is not a sufficient tool for human skill transfer. Consider the question answering example in~\cref{fig:teaser}(a). The LLM can summarize general world knowledge, but its response may not be directly applicable to the user's current circumstance. To bridge this gap, egocentric visual perception provides a novel means to capture the actions and intents as well as the surrounding context of the camera wearer. As shown in~\cref{fig:teaser}(b), recent Visual Large Language Models (VLLMs)~\cite{alayrac2022flamingo,li2023blip2,instructblip,liu2023visual,zhu2023minigpt} can generate instructions based on the egocentric visual input. However, such verbose textual instructions are not the optimal medium for enabling efficient human skill transfer (\eg, via AR devices). Neuroscience studies have revealed that the human brain processes text much more slowly than images~\cite{baskin2007picture}, and that humans can interpret an action from a single static image~\cite{hafri2017neural}. Motivated by these discoveries, we seek to design an image generation architecture that can synthesize an image which not only vividly depicts how an action should be conducted, but also seamlessly aligns with the user's visual perspective. 

Formally, we introduce the novel problem of egocentric action frame generation as shown in \cref{fig:teaser}(c). Given a user query of how to perform a specific action and an egocentric image capturing the moment before the action begins, the goal is to synthesize an egocentric image illustrating the execution of the action in the same egocentric context. In this paper, we address this problem by harnessing diffusion models~\cite{rombach2022high,ho2020denoising}, which have been shown to be powerful tools for image manipulation~\cite{zhang2023inversion,brooks2023instructpix2pix,hertz2022prompt,li2023blipdiff,meng2022sdedit}. There are two major challenges in using diffusion models to generate action frames from an egocentric perspective. First, the action annotations of the existing egocentric datasets~\cite{grauman2022ego4d,damen2022rescaling} are simply composed of a verb and nouns (see word cloud in~\cref{fig:teaser}), and thus lack the necessary details for diffusion models to learn the action state transition and to associate the action with correct objects and body movements. Second, the existing diffusion models are pre-trained mostly on \emph{exocentric} (third-person-view) images, and have limited ability to represent complicated human daily activities from an egocentric perspective. In contrast, our proposed problem requires image generation in the \emph{egocentric} view conditioning on the user prompt of \emph{actions}. The resulting domain gap impedes existing methods from synthesizing faithful and consistent egocentric action frames. 

To address these challenges, we propose to Learn the EGOcentric (LEGO) action frame generation model with visual instruction tuning. First, we introduce a prompt enhancement scheme to generate enriched action descriptions at scale from an instruction tuned VLLM. Second, we incorporate the image and text embeddings from finetuned VLLM as additional conditioning into the diffusion model to narrow the domain gap and improve the controllability of action frame generation. Our experimental results suggest that the enriched action descriptions and our innovative utilization of VLLM embeddings both improve the image generation performance. Our model is able to provide a generated key action frame together with a detailed action descriptions to facilitate human skill transfer from the egocentric perspective. Overall, our contributions can be summarized as follows:
\vspace{-0.2cm}
\begin{itemize}
    \item[$\bullet$] We introduce the novel problem of egocentric action frame generation to facilitate the process of skill transfer and address the challenges of missing action details in prompts and domain gap in existing image diffusion models.
    \item[$\bullet$] We propose a prompt enhancement strategy based on visual instruction tuning to enrich egocentric action descriptions, and demonstrate how the enriched descriptions can help the diffusion model understand the action state transition from the egocentric perspective.
    \item[$\bullet$] We propose a novel approach to incorporate the text and visual embeddings from the VLLM into the latent diffusion model to bridge the domain gap and improve the performance for egocentric action frame generation.
    \item[$\bullet$] We conduct thorough experiments on Ego4D and Epic-Kitchens datasets to validate the superiority of our model over prior approaches. We also showcase the contribution of each component of our model design through ablation studies. We further provide analysis on how the visual instruction tuned embeddings benefit model performance.
\end{itemize}

% -------------------------- Related Work -------------------------------
\section{Related Work}

\textbf{Text-Guided Image Manipulation.} The recent emergence of diffusion models enables text-guided image manipulation including image restoration~\cite{jiang2023autodir}, style transfer~\cite{sun2023sgdiff}, personalized image synthesis~\cite{ruiz2023dreambooth,shi2023instantbooth,wei2023elite}, pose generation~\cite{kawar2023imagic,huang2023kv} and generic image editing~\cite{zhang2023sine,wang2023dynamic,yu2023fisedit,joseph2023iterative,tsaban2023ledits,kim2023user,zhang2023hive,wang2023mdp,mirzaei2023watch,wang2023instructedit,orgad2023editing,li2023blipdiff,wang2023dynamic,goel2023pair,epstein2023diffusion}. SDEdit~\cite{meng2022sdedit} converts the image to the latent space by adding noise through a stochastic differential equation and then denoises the representation for image editing. Rombach \etal~\cite{rombach2022high} further expand SDEdit from the original stroke-based editing to text-based editing. Null-text inversion (NTI)~\cite{mokady2023null} inverts a real image by DDIM~\cite{kawar2022denoising} to yield the diffusion process and then reconstructs the image. The image can then be edited following the same strategies as Prompt-to-Prompt~\cite{hertz2022prompt}. NTI relies on accurate inversion process which can be improved by using coupled transformations~\cite{wallace2023edict} or proximal guidance~\cite{han2024proxedit} and accelerated by a blended guidance strategy~\cite{pan2023effective}. To associate the nouns with correct objects, DIFFEDIT~\cite{couairon2022diffedit} generates a mask to localize the manipulated regions. However, most inversion-based methods require accurate image captions, which are largely unavailable in the existing egocentric dataset. Recently, InstructPix2Pix~\cite{brooks2023instructpix2pix} demonstrates the potential to edit a real image without the inversion process and original captions. However, how to leverage the diffusion model to control the state transition of an action within the egocentric image plane remains unexplored.

\noindent\textbf{Large Language Model for Image Generation.} LLMs~\cite{chowdhery2022palm,brown2020language,raffel2020exploring,zhang2022opt,touvron2023llama,thoppilan2022lamda} and VLLMs~\cite{alayrac2022flamingo,li2023blip2,instructblip,liu2023visual,zhu2023minigpt,zhang2023video,su2023pandagpt,han2023imagebind} have shown their strong capability of understanding complex human instructions. Recently, LLMs and VLLMs are used to guide image generation~\cite{yu2023interactive,llmsmaking,chen2023llava,liu2023acigs,chakrabarty2023learning,zhu2023minigpt,chen2023pixart}. Wen \etal~\cite{wen2023improving} use a pretrained VLLM to pinpoint the misalignment of the text prompt and the synthetic image and then correct it using the diffusion model. Lian \etal~\cite{lian2023llm} propose to generate a layout map using an LLM to improve the understanding of prompts with spatial and numerical reasoning. InstructEdit~\cite{wang2023instructedit} uses BLIP-2~\cite{li2023blip2} to infer the objects that will be edited and then generates a mask with SAM~\cite{kirillov2023segment} for object grounding. A pre-trained LLM can also be used as a controller to connect with various foundation models~\cite{wu2023visual,shen2023hugginggpt,wu2023next}. GILL~\cite{koh2024generating} learns text-relevant image embeddings from VLLM in a few additional tokens for image generation. Importantly, all previous efforts apply the off-the-shelf foundational models directly to their problems without finetuning. In contrast, our method uses visual instruction tuning to improve the image and text embeddings from the VLLM, which narrows the domain gap and thereby facilitates the action frame generation from the egocentric point of view.

\noindent\textbf{Egocentric Vision.} Recent efforts seek to understand human's actions and attentions~\cite{huang2018predicting,huang2020mutual,li2018eye,girdhar2021anticipative,wang2021interactive,sudhakaran2019lsta,kazakos2019epic,lai2022eye,lai2023listen}, modeling hand-object interactions~\cite{liu2020forecasting,ragusa2023stillfast,liu2022joint,goyal2022human,xu2023egopca}, and estimating human body poses~\cite{tome2020selfpose,luo2021dynamics,wang2023scene,li2023ego} from the egocentric perspective. Here, we mainly discuss the most relevant works on egocentric visual-language models and egocentric visual content generation. Lin \etal~\cite{lin2022egocentric} propose the first egocentric video-language pre-training model -- EgoVLP. Pramanick \etal~\cite{pramanick2023egovlpv2} further improve it  by incorporating multi-modal fusion directly into the video and language backbone. Ashutosh \etal~\cite{ashutosh2023hiervl} propose to learn a hierarchical video-language embedding for long egocentric videos. Ramakrishnan \etal~\cite{ramakrishnan2023naq} introduce NaQ, which is a data augmentation strategy to train models for long egocentric video search with natural language queries. In terms of egocentric visual generation, Jia \etal~\cite{jia2022generative} leverage GANs~\cite{goodfellow2020generative} to generate future head motion in hand forecasting task. Zhang \etal~\cite{zhang2017deep} leverage GANs to facilitate future gaze anticipation. Ye \etal~\cite{ye2023affordance} propose the affordance diffusion model that takes in the image of an object and generates possible hand-object interactions in the egocentric view. In this paper, we present the first work that investigates how to leverage VLLMs and diffusion models to generate action state transition on the egocentric image plane.

% -------------------------- Method -------------------------------
\vspace{-0.1cm}
\section{Method}

The problem setting of egocentric action frame generation is illustrated in \cref{fig:teaser}(c). Given an egocentric image frame $\mathcal{X}$ that captures the user's current visual context as well as a user query prompt $\mathcal{P}$ regarding how to perform an action, our goal is to synthesize the action frame $\mathcal{Y}$ that visually depicts how the action should be conducted in the same situation (\ie, keep a consistent background). 

The key insight of our proposed LEGO model is leveraging the strong capability of a VLLM to enhance the diffusion model for egocentric action frame generation. The annotations of existing egocentric datasets do not describe the details of how actions are conducted. As a remedy, we leverage visual instruction tuning to finetune a VLLM that enriches action descriptions based on the egocentric visual prompt. In addition, the existing diffusion-based image manipulation models are limited in understanding egocentric action state transition, due to the domain gap between the exocentric pre-training dataset and the egocentric action dataset for our problem. To bridge this gap, we propose a novel approach that leverages VLLM embeddings to control the state transition of actions and to generate action frames accordingly. We detail the VLLM-based data enrichment pipeline and our model design in the following sections.

\begin{figure}[t]
\centering
\includegraphics[width=0.9\linewidth]{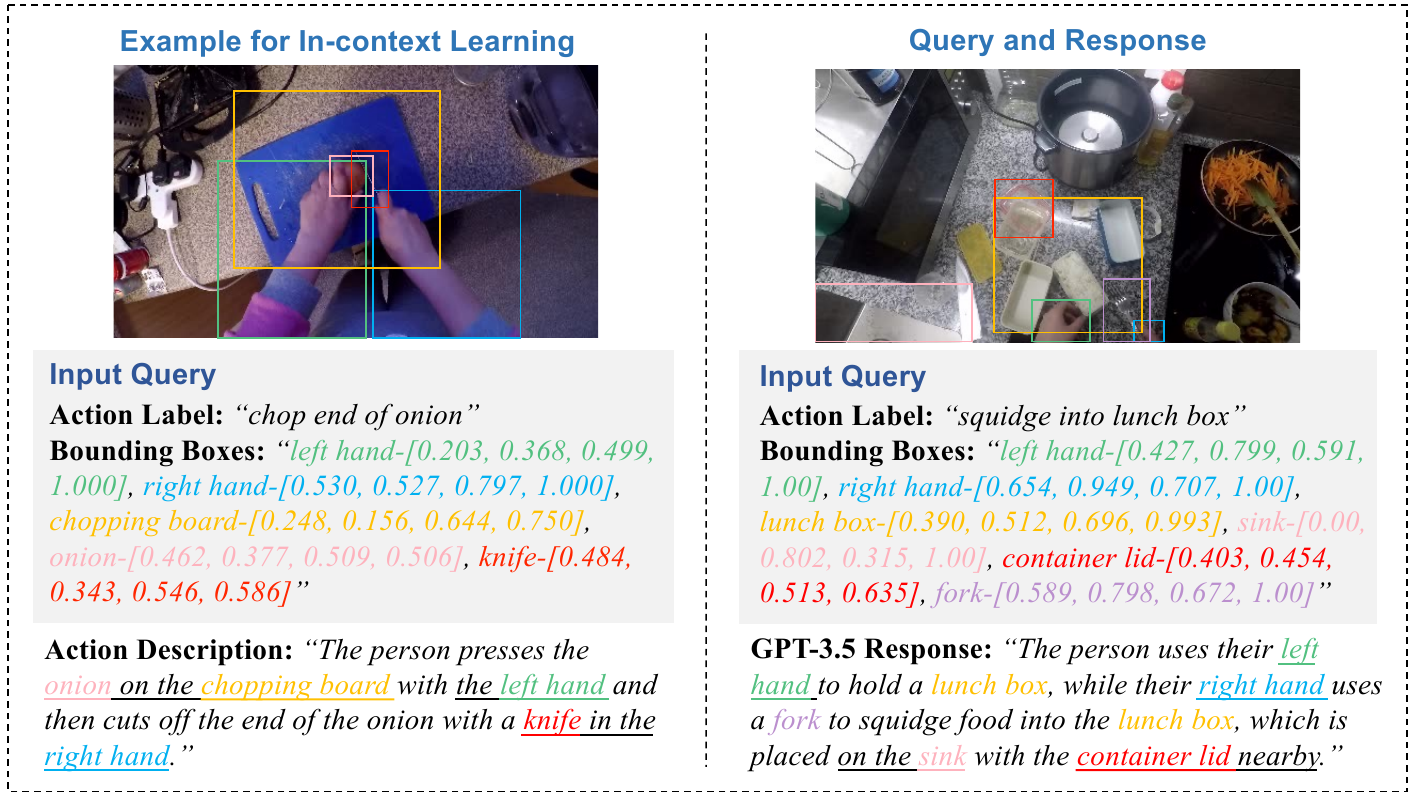}
\vspace{-0.1cm}
\caption{Examples for data curation using GPT-3.5. We provide detailed action descriptions of several example images as well as their action labels and bounding boxes for in-context learning. In addition, we input the action label and bounding boxes of another action as a query. GPT-3.5 is able to generate descriptions with enriched information (highlighted by underlines) in the response.}
\label{fig:curation}
\vspace{-0.3cm}
\end{figure}

\vspace{-0.1cm}
\subsection{Egocentric Visual Instruction Tuning}
\label{sec:instruct_tune}

\textbf{Data Curation for Visual Instruction Tuning.} As shown in ~\cref{fig:curation}, we use GPT-3.5 to generate detailed action descriptions based on an input query composed of a short action label and object bounding boxes that are provided in the existing datasets. First, we prepare several examples of possible inputs along with their expected output descriptions for GPT-3.5 to perform \emph{in-context learning}. These examples cover a diverse set of scenes in the egocentric action dataset. Each example is composed of an action label, a manually annotated action description, and relative spatial information of hands and objects-of-interests represented by bounding box coordinates. GPT-3.5 can learn from the given examples and generate similar detailed action descriptions based on a new input query. The resulting GPT-3.5 curated data is then further used for visual instruction tuning. More details of the prompt are provided in the supplementary.

\begin{figure}[t]
\centering
\includegraphics[width=0.99\linewidth]{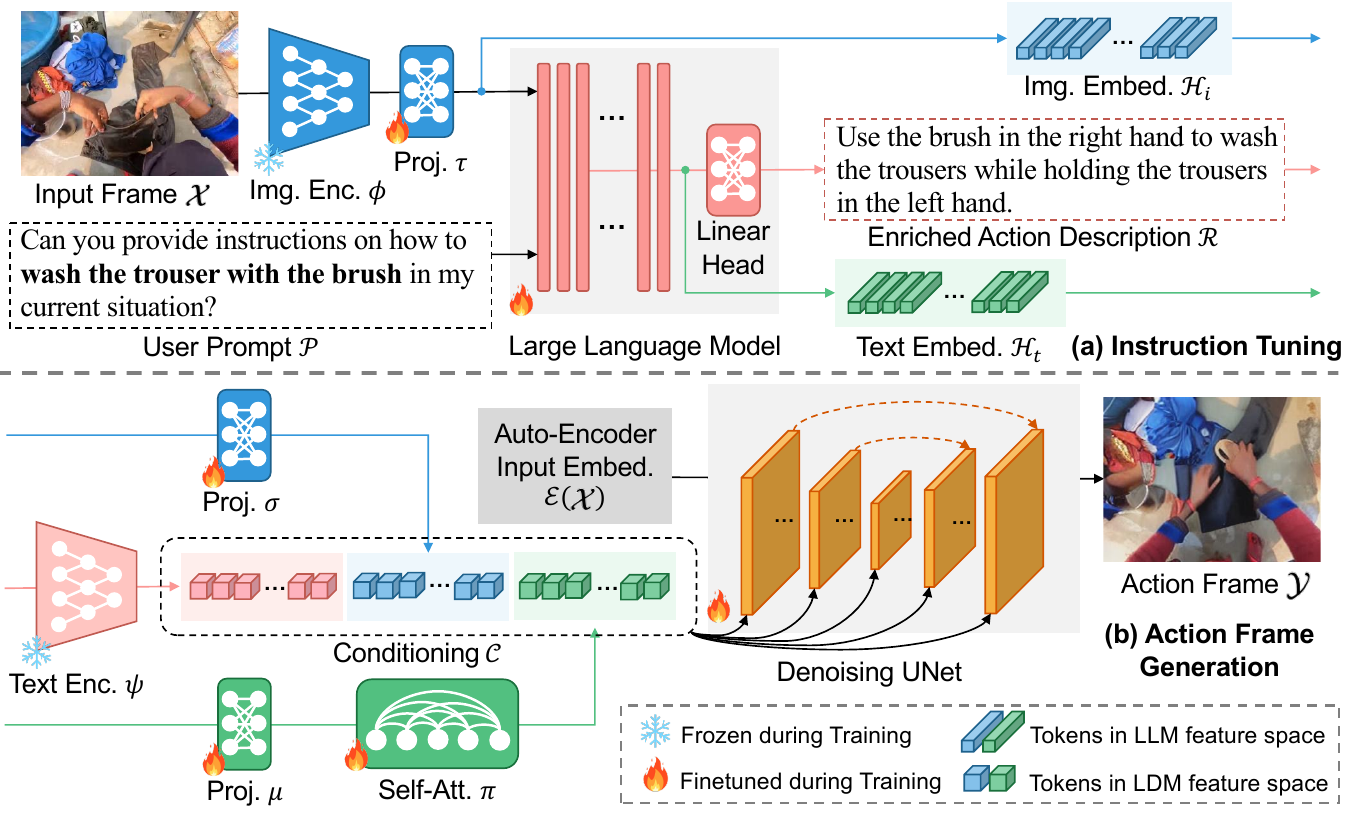}
\vspace{-0.2cm}
\caption{Overview of our proposed LEGO model. We first finetune a visual large language model (VLLM) to generate the enriched action description with visual instruction tuning. We then project image and text embeddings from the finetuned VLLM to the feature space of the latent diffusion model (LDM). Finally, we train the LDM to synthesize the egocentric action frame conditioning on the input frame, enriched action description, as well as the VLLM image and text embeddings.}
\vspace{-0.3cm}
\label{fig:model}
\end{figure}

\noindent\textbf{Visual Instruction Tuning.} We follow the finetuning strategy in prior work~\cite{liu2023visual}, as shown in \cref{fig:model}(a). Specifically, we use the pre-trained CLIP visual encoder~\cite{radford2021learning} $\phi$ to encode the visual representation and then apply a linear projection layer $\tau$ to map the CLIP visual features into the semantic space of the LLM, \ie, $\mathcal{H}_i = \tau(\phi(\mathcal{X}))$. To construct the user prompt $\mathcal{P}$, we insert the action label annotation into a prompt template to create a coherent query that is aligned with the instruction-following nature of an LLM. We then tokenize $\mathcal{P}$, and feed both the prompt text tokens and image tokens $\mathcal{H}_i$ as inputs into the LLM. Finally, the LLM is trained to generate enriched action description (denoted as $\mathcal{R}$) based on the user prompt and image input. 

\noindent\textbf{User Prompt Enrichment at Scale.} Note that the visual instruction tuned VLLM doesn't rely on any object bounding boxes as input. Therefore, we can generate enriched action descriptions for all egocentric action data at scale.

\vspace{-0.3cm}
\subsection{Egocentric Action Frame Generation}
\label{sec:method_image_generation}

We leverage a latent diffusion model (LDM)~\cite{rombach2022high} to synthesize the action frame conditioning on the input frame and the detailed action description $\mathcal{R}$ generated by our finetuned VLLM (see \cref{fig:model}(b)). Following the regular steps in LDMs~\cite{rombach2022high,brooks2023instructpix2pix}, the input image is first encoded into a latent space using a pre-trained auto-encoder $\mathcal{E}$. Then the input to the denoising UNet is a concatentation of the latent input representation $\mathcal{E}(\mathcal{X})$ and a Gaussian noise $\mathcal{G}$. 

Our key innovation is to design the U-Net conditioning component so that the diffusion model can interpret the egocentric actions correctly. To start, we follow~\cite{brooks2023instructpix2pix} and adopt the conventional pre-trained CLIP text encoder $\psi$ to extract a text representation of $\mathcal{R}$, \ie, $\psi(\mathcal{R})\in\mathbb{R}^{N\times D}$ where $N$ is the maximum number of text tokens and $D$ is the number of feature channels. We further leverage the image and text embeddings from the visual instruction tuned VLLM as additional LDM conditioning to alleviate the domain gap issue.

Specifically, we feed the VLLM image embedding $\mathcal{H}_i$ into an extra linear layer $\sigma$ to map it to LDM feature space, \ie, $\sigma(\mathcal{H}_i)\in\mathbb{R}^{M\times D}$, where $M$ is the number of image tokens. Note that $\mathcal{H}_i$ is already projected to the semantic space during visual instruction tuning, and therefore differs from the image embedding $\mathcal{E}(\mathcal{X})$ from the auto-encoder. Moreover, we also extract the text embedding $\mathcal{H}_t$ before the last linear layer of LLM. We adopt a fixed token number $N$ and enforce padding or truncation behavior, as in the CLIP text encoder. The text embedding is then fed to a projection layer $\mu$. In LLM decoder, the response is generated iteratively and each word embedding only conditions on the context ahead of it. To extract the holistic semantic meaning of $\mathcal{H}_t$ in LDM feature space, we further add self-attention layers $\pi$ after the projection, \ie, $\pi(\mu(\mathcal{H}_t))\in\mathbb{R}^{N\times D}$. Thus, the U-Net conditioning can be formulated as:
\begin{equation}
    \mathcal{C} = \Big[\psi(\mathcal{R}), \sigma(\mathcal{H}_i), \pi(\mu(\mathcal{H}_t))\Big] \in\mathbb{R}^{(2N+M)\times D}.
\end{equation}
The conditioning $\mathcal{C}$ is fed into the denoising UNet at multiple layers via the cross-attention mechanism~\cite{rombach2022high}. We assume the intermediate feature of a specific UNet layer is $\mathcal{U}$, which is learned from the UNet input (\ie, the concatenation of input frame representation $\mathcal{E}(\mathcal{X})$ and Gaussian noise $\mathcal{G}$). The cross-attention at this UNet layer can be formulated as:
\begin{equation}
    CrossAtt(Q, K, V) = softmax\left(\frac{QK^T}{\sqrt{D}}\right)\cdot V,
\end{equation}
where $Q=W_Q\cdot\mathcal{U}$, $K=W_K\cdot\mathcal{C}$ and $V=W_V\cdot\mathcal{C}$. Note that $W_Q$, $W_K$ and $W_V$ are learnable matrices. We also adopt the classifier-free guidance following~\cite{brooks2023instructpix2pix} (see supplementary for details). Finally, the UNet output is converted to the image domain by a pre-trained decoder. 

\subsection{Implementation Details}
All parameters of the VLLM are initialized from the pre-trained LLaVA~\cite{liu2023visual} weights. During training, we freeze the CLIP image encoder and finetune the projection layer and LLM with cross-entropy loss for 3 epochs. To improve the diversity of the text prompts, we randomly select the question template from 10 candidates in each iteration. For LDM training, the text encoder, UNet and auto-encoder are initialized with pre-trained weights~\cite{rombach2022high}. The projection layers $\sigma$ and $\mu$ and the self-attention layers $\pi$ are initialized using the Xavier algorithm~\cite{glorot2010understanding}. The text encoder is frozen and the remaining weights are finetuned with L2 regression loss between the predicted noise and real noise for 20,000 iterations. All input and target images are resized to a resolution of  256$\times$256. Please refer to the supplementary for more details of training and inference.

% -------------------------- Experiments -------------------------------
\vspace{-0.1cm}
\section{Experiments}

We first introduce the datasets, data preprocessing, and evaluation metrics used in our experiments. Then we demonstrate the improvement of our model over prior image manipulation approaches in both quantitative evaluation and qualitative visualization. In addition, we ablate our model to show the contribution of each component and the importance of finetuning. We also evaluate the quality of enriched action descriptions generated by VLLM.  Finally, we show the capability of our model in synthesizing various actions with the same image input.

\vspace{-0.1cm}
\subsection{Data and Metrics}
\label{sec:data_metrics}

\textbf{Datasets.} We conduct our experiments on two well-established egocentric action datasets -- Ego4D~\cite{grauman2022ego4d} and Epic-Kitchens-100~\cite{damen2022rescaling}. Both datasets were densely annotated with action starting time $t$ and ending time $\hat{t}$. In our problem setting, we select an egocentric image frame $\delta_i$ seconds before the action begins as the input $\mathcal{X}$, and an image $\delta_o$ seconds after the action begins as the target frame $\mathcal{Y}$. On the Ego4D dataset, based on the annotations of Pre-Condition-15 time (PRE-15)  $t_{pre}$ , and Point-of-No-Return time (PNR) $t_{pnr}$, we set $\delta_i=t-t_{pre}$ and $\delta_o=t_{pnr}-t$. For Epic-Kitchens, PNR and PRE-15 annotations are not available. Instead, we empirically select $\delta_i=0.25$ seconds and $\delta_o=t+\lambda*(\hat{t}-t)$, where $\lambda=0.6$, for our experiments.

\noindent\textbf{Data Preparation for Visual Instruction Tuning}. 
For our dataset curation, we randomly select 20,891 actions with bounding box annotations from the Ego4D training set and 17,922 actions with VISOR~\cite{VISOR2022} mask annotations from the Epic-Kitchens training set. We leverage GPT to produce detailed descriptions of these actions as described in \cref{sec:instruct_tune}. For instruction tuning, we insert the original action label into a prompt template to construct the full user prompt. In order to diversify the prompt structure, we prepare 10 prompt templates and randomly select one for each action at training time.

\noindent\textbf{Data Improvement for Action Frame Generation}. Due to the possible drastic camera motion, the egocentric image frames at $t-\delta_i$ and $t+\delta_o$ may be blurry. As a mitigation, we first calculate aesthetic scores~\cite{aesthetic_score} of the frames at $t-\delta_i$ and $t+\delta_o$ as well as $3$ frames before and after them. The corresponding frames with the highest aesthetic score are used as the input and ground truth of our model. In addition, the egocentric body motion may have huge variance depending on the action type, meaning that the input frame and target frame may look almost identical in more stationary actions (\eg, camera wearer is reading book), or significantly different in more active actions (\eg, outdoor activities with frequent head motion). Such large variances may incur additional barriers for the diffusion model training. Therefore, we calculate the similarity of the input frame and target frame, and we filter out the instances where the similarity is lower than 0.81 or higher than 0.97. With these steps, we ultimately curate 85521/9931 data samples for the train/test sets from Ego4D and 61841/8893 data samples for the train/test sets from Epic-Kitchens.

\noindent\textbf{Metrics.} We adopt image-to-image similarity, image-to-text similarity, and user study as metrics in our experiments. 
\begin{itemize}
    \setlength\itemsep{0.1em}
    \item[$\bullet$] \textbf{Image-to-Image Metrics.} We implement six metrics to evaluate image-to-image similarity. To begin with, we adopt three contrastive learning based metrics including image-to-image (1) EgoVLP score~\cite{lin2022egocentric}, (2) EgoVLP$^+$ score \cite{lin2022egocentric} and (3) CLIP score~\cite{radford2021learning}. EgoVLP is a contrastive video-language pre-training model trained with egocentric videos. Since EgoVLP takes multiple frames as input, we consider two types of inputs: duplicating the output frame as a static input (\ie, EgoVLP score) and combining the input frame with the output frame (\ie, EgoVLP$^+$ score). As as result, EgoVLP$^+$ can effetively measure whether the generated frame can depict the state transition of an action. Importantly, given that EgoVLP is pre-trained on egocentric data and action labels, we consider EgoVLP and EgoVLP$^{+}$ score as the \emph{primary} automatic metrics. In addition, we also report (4) Fréchet Inception Distance (FID)~\cite{heusel2017gans}, (5) Peak Signal-to-Noise Ratio (PSNR) and (6) Learned Perceptual Image Patch Similarity (LPIPS)~\cite{zhang2018unreasonable} (with SqueezeNet~\cite{iandola2016squeezenet} as the encoder) to make a thorough evaluation. Note that instead of measuring similarity of input and output frames as in prior works~\cite{instructblip,wallace2023edict}, in our problem setting, we measure the similarity between the generated action frame and ground truth, which better reflects whether the generation results can illustrate the execution of an action.
    \item[$\bullet$] \textbf{Image-to-Text Metrics.} We find the widely-used image-to-text CLIP score can not align actions with egocentric images due to the domain gap. Similar misalignment problem is also observed in \cite{molad2023dreamix,wang2023zero,du2023learning,stein2023exposing}. In our experiments, we utilize BLIP~\cite{li2022blip} to generate image captions of the output images and then calculate text-to-text similarity using CLIP text encoder (following~\cite{joseph2023iterative}). We implement this metric with two BLIP structures: BLIP-B and BLIP-L. Though this solution may still suffer from the same domain gap issue, it is a more appropriate evaluation metric in our problem setting. See more evidence and discussions in the supplementary.
    \item[$\bullet$] \textbf{User Study.} We also conduct a user study on a subset of test data to further validate the advantage of our model based on human preference. We sample 60 examples from each dataset and present the generated frames from our model as well as the baseline models to raters on Amazon Mechanical Turk (AMT). We also provide the input frames and the corresponding action labels during evaluation. For each instance, we ask the raters to select the image that best aligns with the provided action label while preserving the most contextual information from the input frame. To minimize potential bias, we hire 5 AMT raters to annotate each example thus resulting in 300 samples for each dataset. User study interface is shown in supplementary.
\end{itemize}

\begin{table}[t]
\caption{Comparison with prior image manipulation approaches in image-to-image metrics. $\downarrow$ means a lower score in this metric suggests a better performance. The best results are highlighted with \textbf{boldface}. The {\color{orange} orange} row refers to our LEGO model.}
\vspace{-0.2cm}
\centering
\setlength{\tabcolsep}{0.17cm}
\begin{tabular}{c@{\hspace{0.1cm}}|lcccccc}
\toprule
& Methods & EgoVLP & EgoVLP$^+$ & CLIP  & FID $\downarrow$ & PSNR & LPIPS $\downarrow$ \\
\midrule
\multirow{4}{*}{\rotatebox[origin=c]{90}{Ego4D}}
& ProxEdit~\cite{han2024proxedit}       & 44.51 & 72.68 & 68.17 & 33.01 & 11.88 & 40.90 \\
& SDEdit~\cite{meng2022sdedit}          & 50.07 & 72.90 & 73.35 & 33.35 & 11.81 & 41.60 \\
& IP2P~\cite{brooks2023instructpix2pix} & 62.19 & 78.84 & 78.75 & 24.73 & 12.16 & 37.16 \\
& \cellcolor[HTML]{FAEBD7}LEGO & \cellcolor[HTML]{FAEBD7}\textbf{65.65} & \cellcolor[HTML]{FAEBD7}\textbf{80.44} & \cellcolor[HTML]{FAEBD7}\textbf{80.61} & \cellcolor[HTML]{FAEBD7}\textbf{23.83} & \cellcolor[HTML]{FAEBD7}\textbf{12.29} & \cellcolor[HTML]{FAEBD7}\textbf{36.43} \\
\midrule
\multirow{4}{*}{\rotatebox[origin=c]{90}{E-Kitchens}}  
& ProxEdit~\cite{han2024proxedit}       & 32.27 & 52.77 & 65.80 & 51.35 & 11.06 & 46.35 \\
& SDEdit~\cite{meng2022sdedit}          & 33.84 & 56.80 & 74.76 & 27.41 & 11.30 & 43.33 \\
& IP2P~\cite{brooks2023instructpix2pix} & 42.97 & 61.06 & 77.03 & \textbf{20.64} & 11.23 & 40.82 \\
& \cellcolor[HTML]{FAEBD7}LEGO & \cellcolor[HTML]{FAEBD7}\textbf{45.89} & \cellcolor[HTML]{FAEBD7}\textbf{62.66} & \cellcolor[HTML]{FAEBD7}\textbf{78.63} & \cellcolor[HTML]{FAEBD7}21.57 & \cellcolor[HTML]{FAEBD7}\textbf{11.33} & \cellcolor[HTML]{FAEBD7}\textbf{40.36} \\
\bottomrule
\end{tabular}
\vspace{-0.2cm}
\label{tab:image_image_cmp}
\end{table}

\begin{figure}[t]
\centering
\begin{minipage}{0.5\textwidth}
    \scriptsize
    \tabcaption{Image-to-text metrics of our model and baselines. The best results are highlighted with \textbf{boldface}. The {\color{orange} orange} row refers to our LEGO model performance.}
    \centering
    \begin{tabular}{lcccc}
        \toprule
        \multirow{2}{*}{Methods}  & \multicolumn{2}{c}{Ego4D} & \multicolumn{2}{c}{Epic-Kitchens} \\
        \cmidrule(lr){2-3} \cmidrule(lr){4-5}
        & BLIP-B & BLIP-L & BLIP-B & BLIP-L \\
        \midrule
        ProxEdit~\cite{han2024proxedit}       & 17.73 & 17.35 & 23.65 & 23.39 \\
        SDEdit~\cite{meng2022sdedit}          & 19.80 & 19.74 & 21.51 & 21.30 \\
        IP2P~\cite{brooks2023instructpix2pix} & 20.00 & 20.56 & 25.37 & 26.36 \\
        \rowcolor[HTML]{FAEBD7} LEGO & \textbf{20.38} & \textbf{20.70} & \textbf{26.98} & \textbf{27.41} \\
        \bottomrule
    \end{tabular}
    \label{tab:image_text_cmp}
\end{minipage}%
\begin{minipage}[h]{0.45\linewidth}
    \centering
    \includegraphics[width=0.95\linewidth]{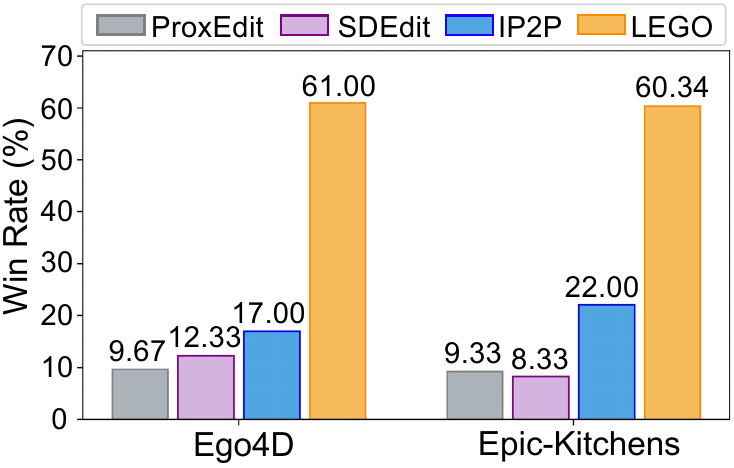}
    \vspace{-0.3cm}
    \caption{User study of our model and baselines. Win rate is the percentage of each model being picked as the best.}
    \label{fig:userstudy}
\end{minipage}
\vspace{-0.4cm}
\end{figure}

\subsection{Comparison with Prior Approaches}
We compare our proposed model with previous state-of-the-art methods for text-guided image manipulation, including ProxEdit~\cite{han2024proxedit}, SDEdit~\cite{meng2022sdedit} and InstructPix2Pix (IP2P)~\cite{brooks2023instructpix2pix}. For a fair comparison, we finetune these baseline methods with the same data used in our experiments. Specifically, we train IP2P in an end-to-end way on the two datasets with existing egocentric action labels as text conditioning. ProxEdit and SDEdit rely on the off-the-shelf latent diffusion model to synthesize edited images and thus can not be trained end to end. Therefore, we first finetune the latent diffusion model with egocentric images and action labels and then use the finetuned latent diffusion model parameters to implement ProxEdit and SDEdit approaches. Please refer to the supplementary for more baseline implementation details and a thorough comparison with prior methods that leverage LLMs for image generation.

In terms of image-to-image metrics in \cref{tab:image_image_cmp}, both ProxEdit and SDEdit perform poorly on this novel problem, suggesting the challenging nature of generating egocentric action frames. IP2P performs much better in almost all metrics by end-to-end training and serves as the strongest baseline model in our experiments. However, our LEGO model consistently outperforms IP2P by a large margin (3.46\%, 1.60\%, 1.86\%, 0.90, 0.13 and 0.73\% respectively) in all six metrics on Ego4D. LEGO also exceeds IP2P notably (2.92\%, 1.60\%, 1.60\%, 0.10 and 0.46\% respectively) in five metrics on Epic-Kitchens. Though, LEGO slightly lags behind IP2P in FID score, it still achieves the second best performance.

\begin{figure}[t]
\centering
\includegraphics[width=0.92\linewidth]{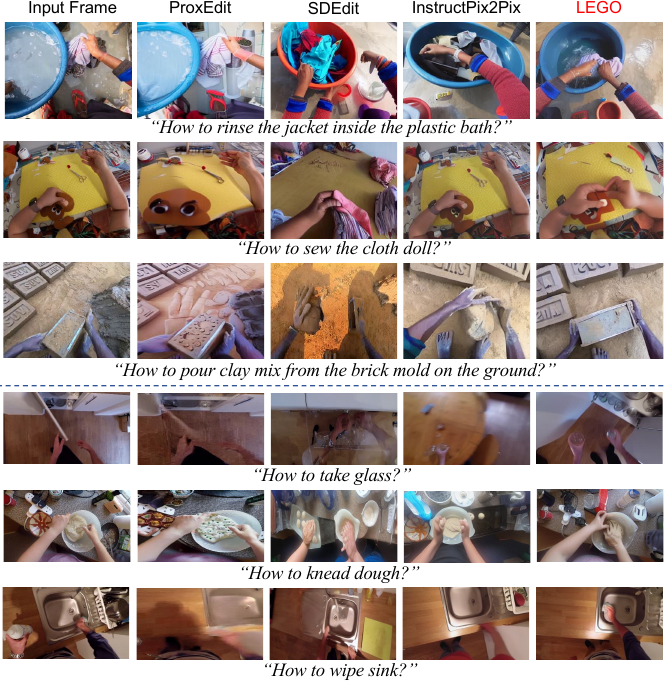}
\vspace{-0.1cm}
\caption{Visualization of the proposed LEGO model and all baselines on Ego4D (the first three rows) and Epic-Kitchens (the last three rows). The action frames generated by LEGO align with the user prompt better and preserve more contexts in input frames.}
\vspace{-0.3cm}
\label{fig:visualization}
\end{figure}

With regard to image-to-text metrics in \cref{tab:image_text_cmp}, LEGO outperforms the strongest baseline (IP2P) by 0.38\% and 0.14\% on Ego4D and by 1.61\% and 1.05\% on Epic-Kitchens. The result suggests the synthetic frames from our model can better align with the action descriptions. However, the performance gain is rather limited. We emphasize that this is because the domain gap still exists for the BLIP model. Besides baseline models, we also measure the image-to-text similarity of input frames and action prompts to validate models' capability of learning action state transition in \emph{semantics}. The BLIP-B/BLIP-L scores are 15.44\%/15.49\% on Ego4D and 20.11\%/20.52\% on Epic-Kitchens, lagging behind all baseline models. The result evidences diffusion models are able to understand actions and edit the input frame \emph{semantically} towards the action prompt.

Due to the limitation of automatic metrics, we additionally implement user study as a complement. We shuffle the order of the results from our model and the baselines while presenting them to the raters. We define the win rate as the percentage of each model being picked as the best out of the total 300 samples. Results are illustrated in \cref{fig:userstudy}. Our model surpasses the best baseline by 44.00\% and 38.34\% respectively on Ego4D and Epic-Kitchens. The prominent gap further validates the superiority of our model.

\subsection{Visualization of Generated Frames}

\begin{figure}[t]
\centering
\includegraphics[width=\linewidth]{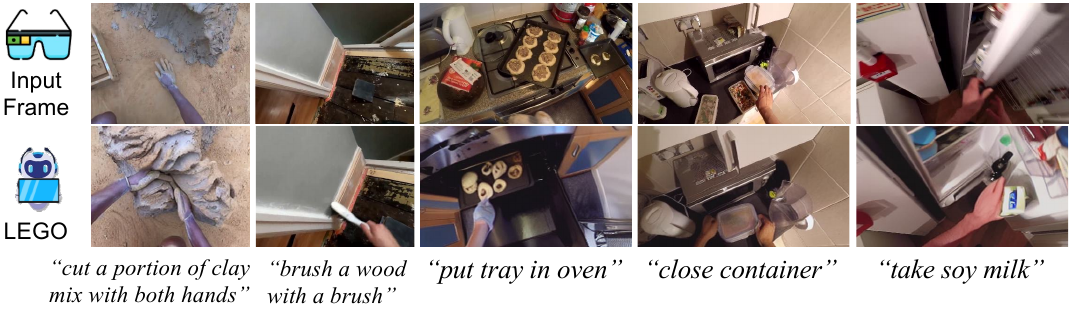}
\vspace{-0.5cm}
\caption{Additional visualization of our proposed model. LEGO is able to synthesize faithful action frames as well as preserve the contexts in various scenarios.}
\vspace{-0.2cm}
\label{fig:extra_visualization}
\end{figure}

We additionally showcase examples of generated image from LEGO and baseline models in ~\cref{fig:visualization}. ProxEdit and SDEdit fail to understand the user prompt and thus generate frames of irrelevant actions (\eg, row3). They may also easily synthesize the action in a different environment (\eg, row2). InstructPix2Pix is able to preserve more contexts but still fails to semantically align with the action in user prompt (\eg, row1). In contrast, LEGO can synthesize action frames that better align with the user prompts and retain more contexts in the background. More examples of LEGO output are presented in \cref{fig:extra_visualization} and supplementary.

\subsection{Ablation Study}
\label{sec:ablation}

We present comprehensive ablation studies to investigate the contribution of enriched action descriptions, the VLLM image embedding and the VLLM text embedding separately. Results are demonstrated in \cref{tab:ablation}. Given the limitation of automatic metrics, we also provide user study as the \emph{primary} metric. Notably, conditioning on enriched action descriptions can moderately improve the model performance, supporting our claim that expanding the action description can facilitate the learning of the state transition during an egocentric action. Moreover, utilizing the image embedding and text embedding from VLLM as additional conditions both improve the model performance by a notable margin because VLLM embeddings can effectively narrow the domain gap. Interestingly, the image embedding leads to larger performance boost on both datasets. These results suggest the VLLM image embedding $\mathcal{H}_i$ incorporates important high-level semantic meanings that are not captured in the auto-encoder image embedding $\mathcal{E}(\mathcal{X})$ or VLLM text emebdding $\mathcal{H}_t$. Finally, our full LEGO model uses both VLLM image and text embeddings (Desc.+Joint Embed.) and thus achieves the best performance on both datasets in all metrics.

\begin{table}[t]
\scriptsize
\caption{Analysis of egocentric action frame generation performance with different conditionings. Joint Embed. refers to incorporating both VLLM image and text embeddings. Similar to \cref{fig:userstudy}, we present win rate as the user study result, \ie, the percentage of each model picked as the best (\% is omitted for simplicity). The best results are highlighted with \textbf{boldface}. The {\color{orange} orange} rows refer to our full LEGO model.}
\vspace{-0.2cm}
\centering
\begin{tabular}{lcccccccc}
\toprule
\multirow{2}{*}{Conditioning} & \multicolumn{4}{c}{Ego4D} & \multicolumn{4}{c}{Epic-Kitchens} \\
\cmidrule(lr){2-5} \cmidrule(lr){6-9}
& User Study & EgoVLP & EgoVLP$^+$ & CLIP & User Study & EgoVLP & EgoVLP$^+$ & CLIP \\
\midrule
Actions Labels       & 5.33  & 62.19 & 78.84 & 78.75 & 7.08  & 42.97 & 61.06 & 77.03 \\
Descriptions         & 13.00 & 62.91 & 79.09 & 79.18 & 12.50 & 43.72 & 61.46 & 77.47\\
Desc. + Img Embed.   & 26.00 & 65.35 & 80.13 & 80.57 & 24.17 & 45.82 & 62.29 & 78.60\\
Desc. + Txt Embed.   & 21.33 & 63.29 & 79.40 & 79.21 & 22.08 & 44.68 & 62.02 & 77.74\\
\rowcolor[HTML]{FAEBD7}Desc. + Joint Embed.  & \textbf{34.34} & \textbf{65.65} & \textbf{80.44} & \textbf{80.61} & \textbf{34.17} & \textbf{45.89} & \textbf{62.66} & \textbf{78.63} \\
\bottomrule
\end{tabular}
\vspace{-0.4cm}
\label{tab:ablation}
\end{table}

\begin{table}[t]
\setlength{\tabcolsep}{0.07cm}
\centering
\caption{Performance of LEGO without finetuning (w/o FT). The best results are highlighted with \textbf{boldface}. The {\color{orange} orange} rows refer to our full LEGO model.}
\vspace{-0.2cm}
\begin{tabular}{lcccccc}
\toprule
\multirow{2}{*}{Conditioning} & \multicolumn{3}{c}{Ego4D} & \multicolumn{3}{c}{Epic-Kitchens} \\
\cmidrule(lr){2-4} \cmidrule(lr){5-7}
& EgoVLP & EgoVLP$^+$ & CLIP & EgoVLP & EgoVLP$^+$ & CLIP \\
\midrule
Descriptions                  & 62.91 & 79.09 & 79.18 & 43.72 & 61.46 & 77.47 \\
Desc.+Joint Embed.(w/o FT)    & 64.57 & 79.72 & 80.23 & 44.74 & 61.87 & 78.35 \\
\rowcolor[HTML]{FAEBD7} Desc.+Joint Embed.(w/ FT) & \textbf{65.65} & \textbf{80.44} & \textbf{80.61} & \textbf{45.89} & \textbf{62.66} & \textbf{78.63} \\
\bottomrule
\end{tabular}
\vspace{-0.3cm}
\label{tab:vllm_ft}
\end{table}

\subsection{Analysis of Visual Instruction Tuning}
\label{sec:exp_instruct_tuning}
First, we assess the quality of enriched action descriptions from the visual instruction tuned VLLM by user study. For each sample in user study, we ask the raters to select whether the description aligns with the input and target frames (see supplementary for more details of user study setting and interface). The percentage of samples with aligned frames and action descriptions is \textbf{87\%} on Ego4D and \textbf{84\%} on Epic-Kitchens. The high alignment percentage suggests the visual instruction tuned VLLM can effectively expand action labels with details captured in the input frame. We additionally implement the same user study for \emph{off-the-shelf} VLLM (\ie, without finetuning). The percentage of alignment is just \textbf{27\%} on Ego4D and \textbf{30\%} on Epic-Kitchens with hallucination existing in \textbf{92\%} of unaligned instances. The notable drop supports our argument that visual instruction tuning is a critical step for high-quality prompt enrichment.

In addition, we further investigate whether the visual instruction tuned embeddings can contribute more to the diffusion model than the off-the-self counterpart, which has not been well studied in prior work. As shown in~\cref{tab:vllm_ft}, without any finetuning, the image and text embeddings from VLLM can still improve the baseline model (Descriptions). However finetuned embeddings yield much larger improvement (\eg, 1.08\% and 1.15\% in EgoVLP score) on both datasets. The result suggests that visual instruction tuning is a necessary step to learn semantic image and text embeddings that are more aligned with the egocentric input frame and action prompt, which thus narrows the domain gap and greatly boosts the performance of the latent diffusion model.

\vspace{-0.2cm}
\subsection{Generation of Various Actions with the Same Contexts}

\begin{figure}[t]
\centering
\includegraphics[width=\linewidth]{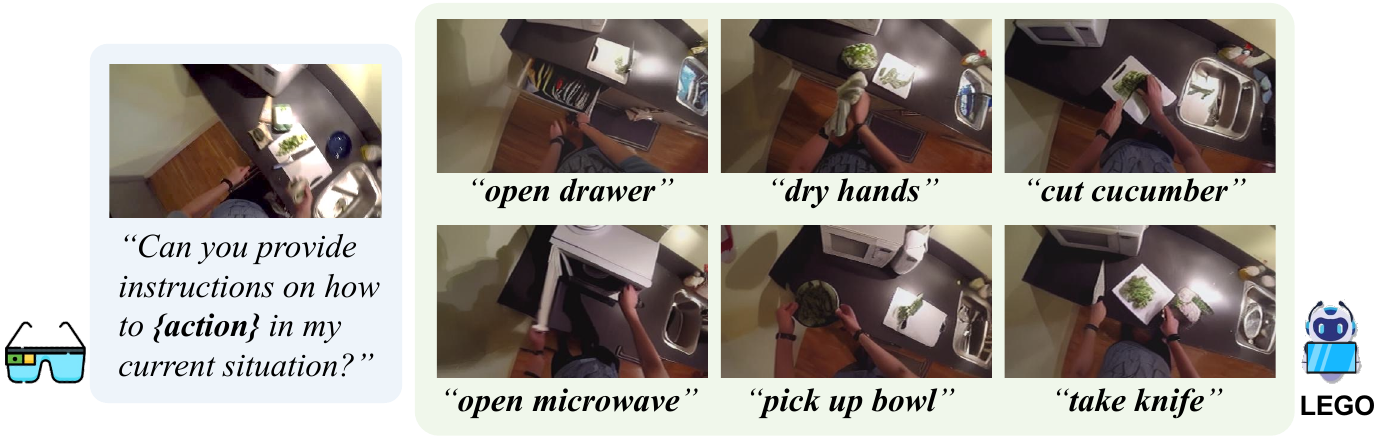}
\caption{Visualization of generating various actions from the same input frame. The first action (\textit{"open drawer"}) is the existing label for this example in the dataset. We fill another five actions into the user query and synthesize the action frames using our model. All generated frames align well with the actions and preserve most contexts.}
\vspace{-0.2cm}
\label{fig:new_actions}
\end{figure}

In addition to generating action frames based on the pre-defined action labels in our datasets, we validate the generative nature of our model \ie, whether LEGO is able to synthesize the correct action frames conditioning on the same contexts yet different actions (novel image-action pairs). Results are illustrated in \cref{fig:new_actions}. We feed different actions with the same input frame to our model. The synthesized frames correctly depict the execution of the actions in the same environment. This result further indicates that our model can understand the action state transition and generalize to different user queries.

\section{Conclusion}

In this paper, we introduce the novel problem of egocentric action frame generation. We also propose a novel model --- LEGO that leverages visual instruction tuning and a diffusion model to address this problem. Our key intuition is to use visual instruction tuning to generate informative responses that depict the execution of the egocentric action, and then design the conditioning for the denoising U-Net to exploit the image and text embeddings from a visual instruction tuned VLLM. Our experiments on two large-scale egocentric action datasets validate the advantage of the proposed approach as well as the contribution of each model component. We believe our work provides an important step in understanding the action state transition and controllability of diffusion models, and suggests future research directions in egocentric AI systems, action state transition, image generation and human skill transfer.

% ---- Bibliography ----
%
% BibTeX users should specify bibliography style 'splncs04'.
% References will then be sorted and formatted in the correct style.
%
\bibliographystyle{splncs04}
\bibliography{egbib}

% ---- Supplementary ----

\clearpage
\appendix
\setcounter{section}{0}

\begin{center}
    \textbf{\Large LEGO: \underline{L}earning \underline{EGO}centric Action Frame Generation via Visual Instruction Tuning}
    \\ [0.8cm]
    {\Large Supplementary Material}
    \\ [1.2cm]
\end{center}

This is the supplementary material for the paper "LEGO: \underline{L}earning \underline{EGO}centric Action Frame Generation via Visual Instruction Tuning". We organize the content as follows:
\\

\noindent\textbf{\hyperref[sec:cmp_third_person]{A} -- Comparison of Egocentric and Exocentric Views} \\[0.2cm]
\noindent\textbf{\hyperref[sec:domain_gap]{B} -- Domain Gap Between Existing Diffusion Models and Our Problem} \\ [0.2cm]
\noindent\textbf{\hyperref[sec:cmp_prior_methods]{C} -- Comparison with Prior Image Generation Models Using LLMs} \\ [0.2cm]
\noindent\textbf{\hyperref[sec:im2txt]{D} -- Analysis of Image-to-Text Metrics} \\ [0.2cm]
\textbf{\hyperref[sec:result]{E} -- Additional Experiment Results} \\ [0.1cm]
\indent\hyperref[sec:transit_time]{E.1} -- Performance at Different Transition Time \\ [0.1cm]
\indent\hyperref[sec:scalability]{E.2} -- Effect of Dataset Scaleup \\ [0.1cm]
\indent\hyperref[sec:visualization]{E.3} -- Additional Visualization \\ [0.2cm]
\textbf{\hyperref[sec:implementation]{F} -- More Implementation Details} \\ [0.1cm]
\indent\hyperref[sec:curation]{F.1} -- Prompt and Examples for Data Curation \\ [0.1cm]
\indent\hyperref[sec:preprocessing]{F.2} -- Examples of Data Preprocessing \\ [0.1cm]
\indent\hyperref[sec:training_details]{F.3} -- Training Details for Instruction Tuning and Action Frame Generation \\ [0.1cm]
\indent\hyperref[sec:classifier_free_guidance]{F.4} -- Details about Classifier-free Guidance \\ [0.1cm]
\indent\hyperref[sec:baseline_models]{F.5} -- Implementation Details of the Baseline Models \\ [0.1cm]
\indent\hyperref[sec:user_study]{F.6} -- Details and Interfaces for User Study \\ [0.2cm]
\textbf{\hyperref[sec:future]{G} -- Limitation and Future Work} \\ [0.2cm]
\textbf{\hyperref[sec:release]{H} -- Code and Data Release}

\renewcommand\thesection{\Alph {section}}
\renewcommand\thesubsection{\thesection.\arabic{subsection}}

\section{Comparison of Egocentric and Exocentric Views}
\label{sec:cmp_third_person}
\begin{figure}[t]
\centering
\includegraphics[width=\linewidth]{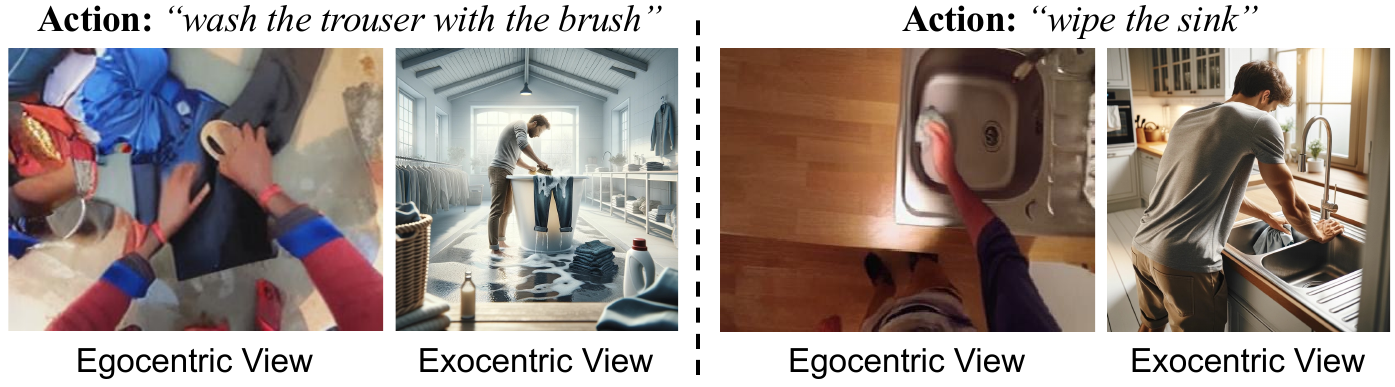}
\caption{Comparison of action frames generated in the egocentric and exocentric views. The egocentric images are synthesized by our model, and the exocentric images are synthesized by an off-the-shelf text-to-image generation model.}
\label{fig:cmp_third_view}
\end{figure}

Our proposed problem and model mainly focus on action frame generation in the egocentric view rather than exocentric view (\ie, third-person view). The examples for the comparison of the two views are illustrated in \cref{fig:cmp_third_view}. The instructional images generated in the exocentric view may be far from the camera. The important details of hand-object interactions are unclear and occluded, thus making it hard for users to follow. In contrast, the action frames generated in the egocentric view by our model exactly match the user's viewpoint and clearly capture important details for action execution. In addition, it's more feasible for users to take a picture that captures the current contexts (\ie, the input frame to our model) from the egocentric perspective than exocentric perspective, especially when the user is working alone. The egocentric image can be obtained even more easily by using a wearable device. These advantages further motivate us to focus on the egocentric view for action frame generation.

\begin{figure}[t]
\centering
\includegraphics[width=\linewidth]{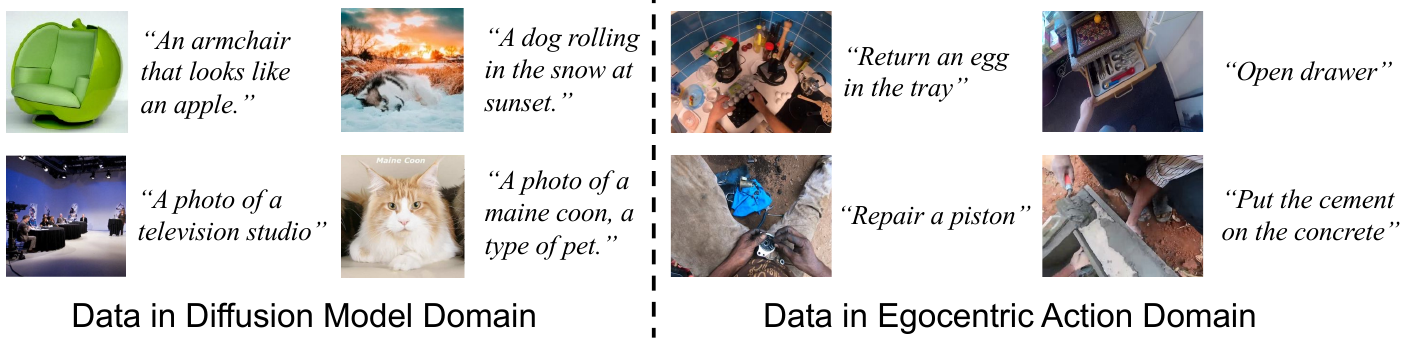}
\caption{Demonstration of the domain gap of off-the-shelf diffusion models and our problem. The existing diffusion models are pre-trained with exocentric images with prompts mainly about objects, while our problem requires the capability of action generation in the egocentric view.}
\label{fig:domain_gap}
\vspace{-0.2cm}
\end{figure}

\section{Domain Gap Between Existing Diffusion Models and Our Problem}
\label{sec:domain_gap}
In the egocentric action frame generation problem, one challenge is the domain gap of the existing diffusion models and our problem (elaborated in \cref{sec:intro} of the main paper). In this section, we show some specific examples to provide more insights. In \cref{fig:domain_gap}, we present a few image-text pairs for text-to-image diffusion model training (left) and data samples for egocentric action frame generation (right). The diffusion model training data is mostly captured from the exocentric perspective and the prompts are about the objects rather than human daily activities. Though we can finetune the model with egocentric data, such a big domain gap still limits the performance of existing diffusion models applied to our problem. To bridge this gap, our proposed LEGO model incorporates the embeddings from a visual instruction tuned LLM as additional conditioning, which enables the diffusion model to learn action state transition more effectively from an egocentric perspective.

\begin{table}[t]
\setlength{\tabcolsep}{1.2pt}
\renewcommand\arraystretch{2}
\scriptsize
\centering
\caption{Comparison with prior text-to-image generation (T2I) and image editing methods that incorporate large language models. Please see \cref{sec:cmp_prior_methods} for more discussions.}
\renewcommand\arraystretch{0.9}
\begin{tabular}{lcccccc}
\toprule
\multirow{2}{*}{Methods}           & Domain & Source of LLM  & How to            & LLM    & Edited      & Main \\
                                   & Gap    & Parameters     & use LLM           & Embed. & Content     & Task \\
\midrule
Chen \etal~\cite{chen2023pixart}   & No     & off-the-shelf  & Auto-label        & w/o    & N/A         & T2I \\
Liu \etal~\cite{liu2023acigs}      & No     & off-the-shelf  & Auto-label        & w/o    & N/A         & T2I \\
Lian \etal~\cite{lian2023llm}      & No     & off-the-shelf  & Enrich Prompt     & w/o    & N/A         & T2I \\
Yu \etal~\cite{yu2023interactive}  & No     & off-the-shelf  & Enrich Prompt     & w/o    & N/A         & T2I \\
Wu \etal~\cite{wu2023next}         & No     & off-the-shelf  & Controller        & w/o    & N/A         & T2I \\ 
Wen \etal~\cite{wen2023improving}  & No     & off-the-shelf  & Multi-round Refine& w/o    & N/A         & T2I \\
Wu \etal~\cite{wu2023visual}       & No     & off-the-shelf  & Multi-round Refine& w/o    & N/A         & T2I \\
\midrule
Chakrabarty \etal~\cite{chakrabarty2023learning}  & No     & off-the-shelf  & Controller        & w/o    & Obj.\&Style & Edit \\
Koh \etal~\cite{koh2024generating}  & No     & off-the-shelf  & Controller        & w/o    & Obj.\&Style & Edit \\
Wang \etal~\cite{wang2023instructedit} & No     & off-the-shelf  & Controller        & w/o    & Object      & Edit \\
Chen \etal~\cite{chen2023llava}       & No     & off-the-shelf  & Multi-round Refine& w/o    & Object      & Edit \\
\midrule
\textbf{LEGO (Ours)}                  & \textbf{Yes} & \textbf{Instruction-tuned} & \textbf{Enrich Prompt} & \textbf{w/} & \textbf{Action} & \textbf{Edit} \\
\bottomrule
\end{tabular}
\label{tab:cmp_prior_methods}
\end{table}

\section{Comparison with Prior Image Generation Models Using LLMs}
\label{sec:cmp_prior_methods}
We compare our model with prior image generation and image editing models that utilize LLMs to improve the performance (see \cref{tab:cmp_prior_methods}). Our model differs from prior methods mainly in four aspects: (1) All prior models are designed for exocentric image generation, while our model is proposed for image generation in the egocentric view, which is still understudied. (2) Prior models are trained with data that doesn't have the domain gap with diffusion model pre-training data. Therefore, prior models can directly use the \emph{off-the-shelf} LLM parameters (\ie, without finetuning) for their tasks. In contrast, our model finetunes the LLM by visual instruction tuning to narrow the domain gap. (3) We also innovatively incorporate LLM embeddings into the diffusion model to boost image generation performance, which has not been investigated in prior work. (4) Prior image editing methods focus on local object manipulation and global style transfer. In our work, we focus on generating images of actions conducted in the same contexts as input, which has not been studied in prior image editing methods. These differences consolidate our contributions and thus notably distinguish our model from prior work.

\section{Analysis of Image-to-Text Metrics}
\label{sec:im2txt}
In \cref{tab:image_text_metric}, we report the image-to-text CLIP score of InstructPix2Pix (IP2P)~\cite{brooks2023instructpix2pix} baseline and the ground truth. Ideally, the CLIP score between the ground truth and the text prompt should serve as a performance uppperbound (UB). However, the image-to-text CLIP score of upperbound is very close to the baseline on Ego4D, and even lower than the baseline on Epic-Kitchens. It suggests the CLIP model fails to align action descriptions with corresponding egocentric images in semantics, thus resulting in a quick saturation in CLIP score. In our experiments, we use BLIP to caption the generated image and measure the text-to-text similarity of captions and action descriptions (following~\cite{joseph2023iterative}). The two metrics (BLIP-B and BLIP-L) that use two different BLIP structures both result in larger gap between the baseline model and the upperbound (3.68\%/2.96\% vs. 0.85\% and 1.61\%/1.44\% vs.$-$1.08\%). Therefore, we adopt BLIP based metrics and user study to measure image-text alignment. Note that, our model still performs on-par or slightly better than IP2P in image-to-text CLIP score, and exceeds IP2P notably when using BLIP based metrics and user study (see \cref{tab:image_text_cmp} and \cref{fig:userstudy} in the main paper).

\begin{table}[t]
\setlength{\tabcolsep}{0.15cm}
\centering
\caption{Image-to-text metrics of the baseline model and uppperbound. The \textcolor{gray}{gray} row shows the gap of IP2P to the upperbound. The upperbound measured by CLIP score is comparable with or even lower than the baseline model (highlighted by \textcolor{red}{red}).}
\begin{tabular}{lcccccc}
\toprule
\multirow{2}{*}{Methods}  & \multicolumn{3}{c}{Ego4D} & \multicolumn{3}{c}{Epic-Kitchens} \\
\cmidrule(lr){2-4} \cmidrule(lr){5-7}
& CLIP & BLIP-B & BLIP-L & CLIP & BLIP-B & BLIP-L \\
\midrule
IP2P~\cite{brooks2023instructpix2pix} & 20.53 & 20.00 & 20.56 & 21.68 & 25.37 & 26.36 \\
UpperBound (UB)                       & 21.38 & 23.68 & 23.52 & 20.60 & 26.98 & 27.80 \\
\rowcolor[HTML]{F0F0F0} $\Delta=$ UB$-$IP2P   & 0.85 & 3.68 & 2.96 & \textcolor{red}{-1.08} & 1.61 & 1.44 \\
\bottomrule
\end{tabular}
\label{tab:image_text_metric}
\end{table}

\begin{table}[t]
\setlength{\tabcolsep}{0.15cm}
\renewcommand\arraystretch{1.2}
\caption{Comparison of LEGO trained with single dataset and both datasets (denoted as \textit{scaleup}). $\downarrow$ means a lower score in this metric suggests a better performance. The better results are highlighted with \textbf{boldface}. The performance of LEGO model can be effectively improved by involving more training data.}
\centering
\begin{tabular}{c|lcccccc}
\toprule
& Methods & EgoVLP & EgoVLP$^+$ & CLIP  & FID $\downarrow$ & PSNR & LPIPS $\downarrow$ \\
\midrule
\multirow{2}{*}{\rotatebox[origin=c]{90}{Ego4D}}
& LEGO            & 65.65 & 80.44 & 80.61 & 23.83 & 12.29 & 36.43 \\
& LEGO (scaleup)  & \textbf{66.32} & \textbf{80.77} & \textbf{80.90} & \textbf{23.56} & \textbf{12.30} & \textbf{36.33} \\
\midrule
\multirow{2}{*}{\rotatebox[origin=c]{90}{EK}}  
& LEGO            & 45.89 & 62.66 & 78.63 & 21.57 & 11.33 & 40.36 \\
& LEGO (scaleup)  & \textbf{47.46} & \textbf{63.51} & \textbf{78.90} & \textbf{19.64} & \textbf{11.40} & \textbf{39.88} \\
\bottomrule
\end{tabular}
\label{tab:scalability}
\end{table}

\begin{figure}[t]
\centering
\includegraphics[width=0.8\linewidth]{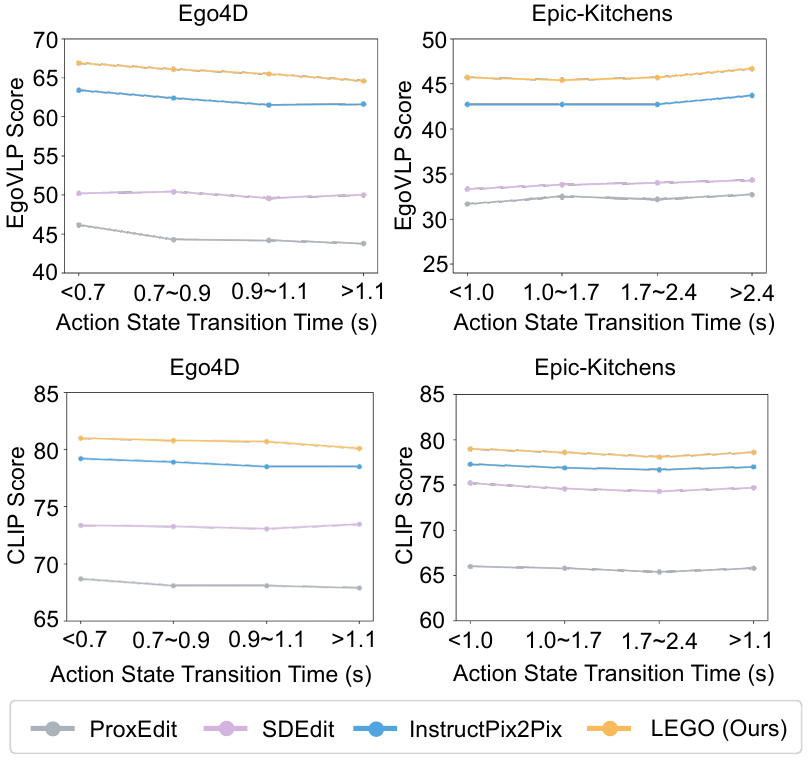}
\caption{Comparison with baselines at different action transition time. Our model outperforms all baselines across all transition time.}
\label{fig:curve}
\vspace{-0.3cm}
\end{figure}

\section{Additional Experiment Results}
\label{sec:result}

\subsection{Performance at Different Transition Time}
\label{sec:transit_time}
As explained in \cref{sec:data_metrics} of main paper, for an action beginning at $t$, we select the frame at $t-\delta_i$ as input and the frame at $t+\delta_o$ as target. We divide the test data into four bins according to the action state transition time from input frame to target frame $\delta=\delta_i+\delta_o$. We establish the threshold for each bin to ensure that the quantity of data samples in each bin is relatively similar. The performance of LEGO and baselines at different transition time is demonstrated in \cref{fig:curve}. The flat curves suggest the egocentric action frame generation problem is equally challenging regardless of transition time. Our model still surpasses all baselines by a notable margin across all transition time. This fine-grained result further validates the superiority of our approach.

\subsection{Effect of Dataset Scaleup}
\label{sec:scalability}
We further analyze how the dataset scale affects our model performance by merging the training set of Ego4D and Epic-Kitchens. The training strategy and other hyper-parameters remain identical to separate training on each dataset (described in \cref{sec:training_details}). We demonstrate the results in \cref{tab:scalability}. The performance of our model is further boosted by leveraging more training data (\ie, scaleup). Notably, the gains on Epic-Kitchens are more prominent than gains on Ego4D (\eg, 1.57\% vs. 0.67\% on EgoVLP score, 1.93 vs. 0.27 on FID, \etc). The possible reason is that Ego4D dataset has more training data covering more diverse scenarios and actions. Hence, it can greatly compensate for the low diversity of Epic-Kitchens dataset after merging them. The improvement on two datasets suggests our model can be effectively boosted by scaling up the training data.

\begin{figure}[t]
\centering
\includegraphics[width=\linewidth]{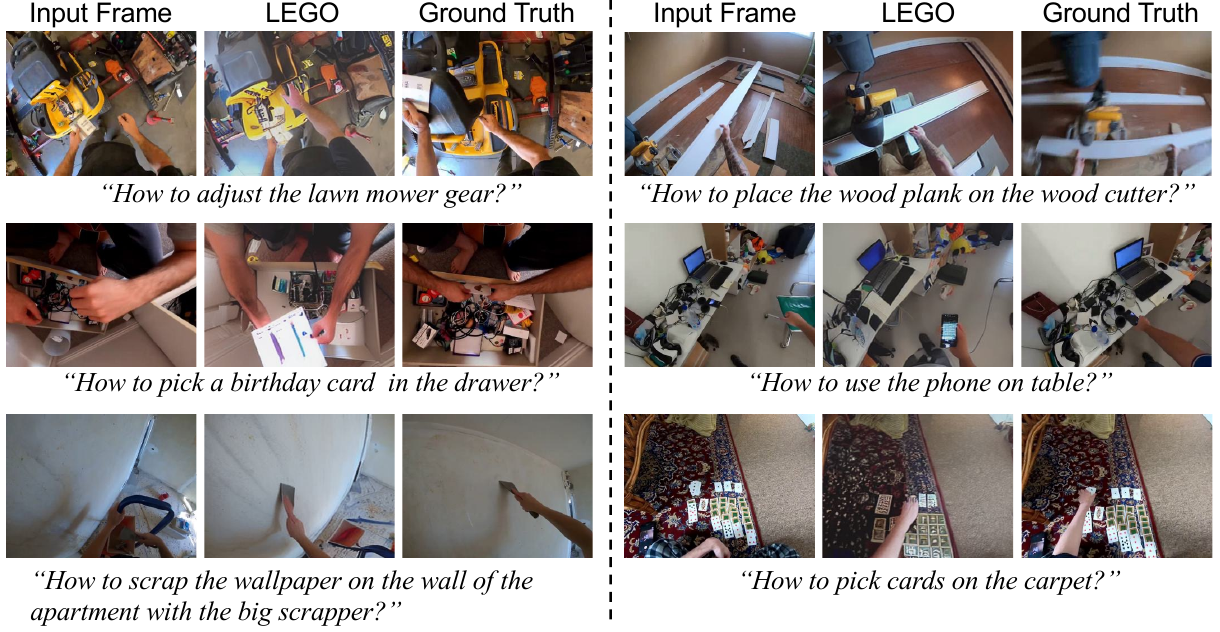}
\caption{Additional visualization of LEGO output on Ego4D as well as the ground truth. Ground truth shows how each action is actually conducted in the real world.}
\label{fig:visualization_ego4d}
\end{figure}

\begin{figure}[t]
\centering
\includegraphics[width=\linewidth]{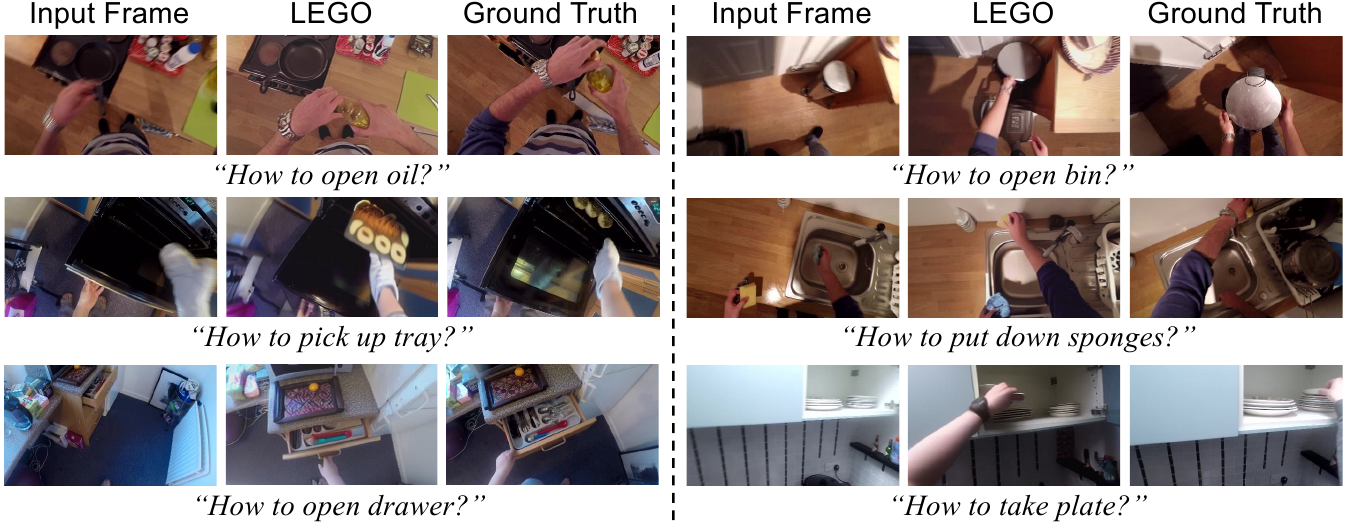}
\caption{Additional visualization of LEGO output on Epic-Kitchens as well as the ground truth. Ground truth shows how each action is actually conducted in the real world.}
\label{fig:visualization_kitchens}
\vspace{-0.2cm}
\end{figure}

\begin{figure}[t]
\centering
\includegraphics[width=\linewidth]{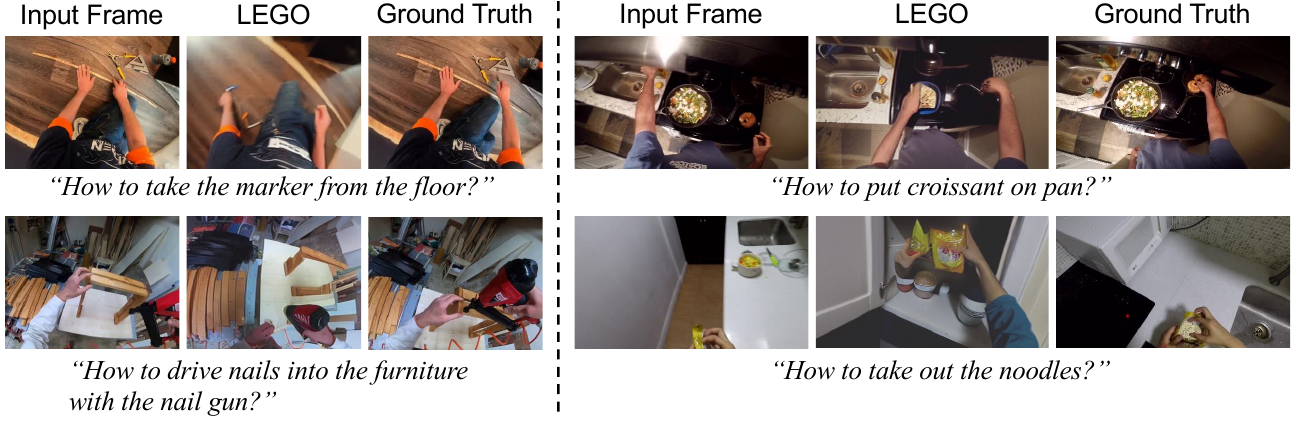}
\caption{Failure cases of our model in Ego4D (on the left of the dash line) and Epic-Kitchens (on the right of the dash line). Please refer to \cref{sec:visualization} for more analysis.}
\label{fig:failure_cases}
\vspace{-0.2cm}
\end{figure}

\subsection{Additional Visualization}
\label{sec:visualization}

We demonstrate more results of our model and the corresponding ground truth on Ego4D (see \cref{fig:visualization_ego4d}) and Epic-Kitchens (see \cref{fig:visualization_kitchens}). The generated frames are well aligned with the user query and the ground truth in these examples. To better understand the limitation of our model, we also illustrate some failure cases in \cref{fig:failure_cases}. Our approach may fail to associate the action with the correct objects when the objects are not distinct enough in the egocentric perspective, \eg, the \textit{marker} and \textit{croissant} in the first row of failure cases. In addition, generating the action frame in some scenarios needs more contexts than a static input frame. For example, the model fails to understand which object is the furniture and incorrectly drives the nails into the wood under it (\ie, the second failure case of Ego4D). It also lacks the context that the user already holds a bag of noodles, so it synthesizes a frame of taking out the noodles from a cupboard (\ie, the second failure case of Epic-Kitchens). These weaknesses can inspire more future studies in action understanding and egocentric action frame generation. Please refer to \cref{sec:future} for more discussions.

\section{More Implementation Details}
\label{sec:implementation}

\begin{figure}[t]
\centering
\includegraphics[width=\linewidth]{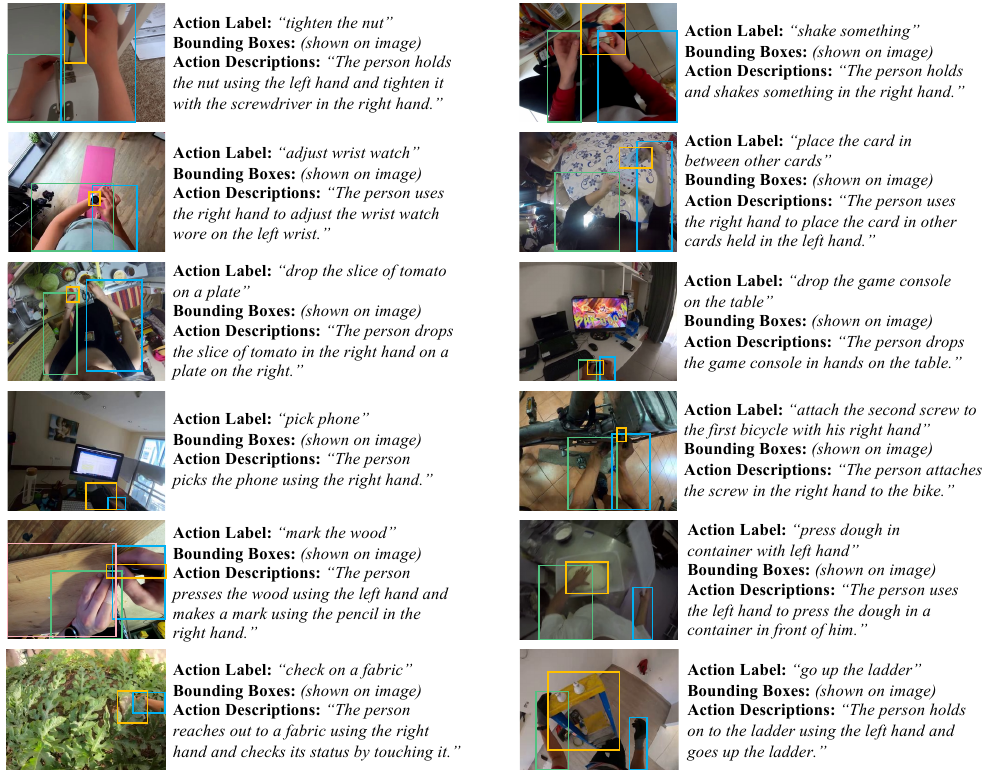}
\caption{All Ego4D examples used for data curation with GPT-3.5 via in-context learning. For simplicity, the bounding boxes are only shown on images. We input the coordinates of bounding boxes to GPT-3.5 in practice.}
\label{fig:curation_ego4d}
\end{figure}

\begin{figure}[t]
\centering
\includegraphics[width=\linewidth]{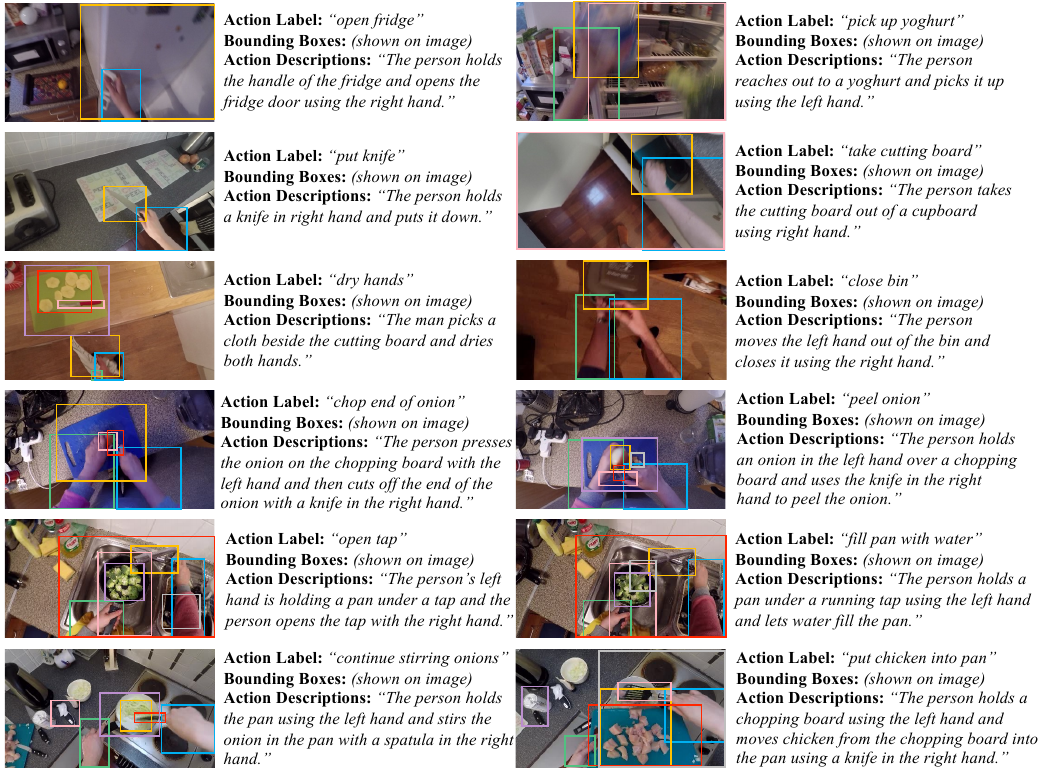}
\caption{All Epic-Kitchens examples used for data curation with GPT-3.5 via in-context learning. For simplicity, the bounding boxes are only shown on images. We input the coordinates of bounding boxes to GPT-3.5 in practice.}
\label{fig:curation_epickitchens}
\end{figure}

\begin{figure}[t]
\centering
\includegraphics[width=\linewidth]{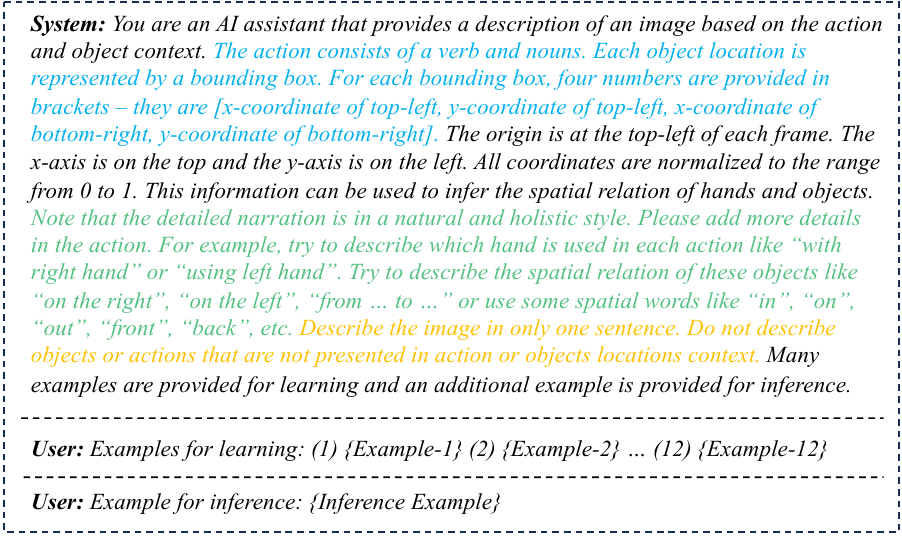}
\vspace{-0.2cm}
\caption{The structure of the prompt sent to GPT-3.5. We specify the composition of the input query (highlighted in \textcolor{cyan}{blue}). Then we articulate the requirements for action enrichment (highlighted in \textcolor{ForestGreen}{green}) and extra demands (highlighted in \textcolor{Dandelion}{yellow}). Example-1 to Example-12 consist of the action label, object bounding boxes and manual annotation of detailed action descriptions. The inference example consists of just action label and object bounding boxes.}
\label{fig:prompt}
\vspace{-0.1cm}
\end{figure}

\subsection{Prompt and Examples for Data Curation}
\label{sec:curation}
In data curation, we randomly select 12 examples from each datasets covering diverse scenarios. All examples are shown in \cref{fig:curation_ego4d} and \cref{fig:curation_epickitchens}. We also clarify our requirements in the prompt sent to GPT-3.5. The complete prompt is shown in \cref{fig:prompt}. We specify the composition of input query, the expected detailed information and extra requirements for the output in the system information. Then we fill the examples for in-context learning and the example for inference in the prompt and input it to GPT-3.5 for data curation.

\begin{figure}[t]
\centering
\includegraphics[width=0.9\linewidth]{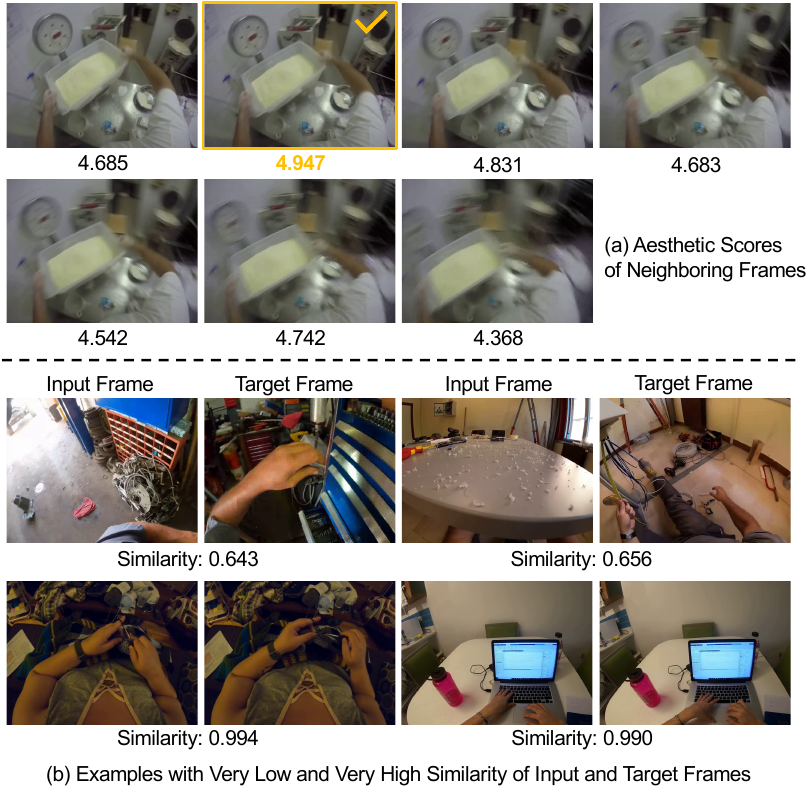}
\caption{Data preprocessing in our work. (a) The fame with the highest aesthetic score (highlighted) is less blurry and then used as the target frame of this action. (b) The actions with too low ($<$0.81) or too high similarity ($>$0.97) between input and target frames are filtered out from the datasets.}
\label{fig:preprocessing}
\vspace{-0.2cm}
\end{figure}

\subsection{Examples of Data Preprocessing}
\label{sec:preprocessing}
For each input frame or target frame, we calculate aesthetic scores~\cite{aesthetic_score} of the current frame as well as 3 frames before and after (7 frames in total). As demonstrated in \cref{fig:preprocessing}(a), the frame of the highest aesthetic score usually has the best image quality. We then use this frame as input or target frame. We also calculate the similarity of input and target frame for each action. We empirically filter out data whose similarity is lower than 0.81 or higher than 0.97. Some examples of abandoned data are shown in \cref{fig:preprocessing}(b). A very low similarity usually indicates a big change in the background due to the head motion. A very high similarity implies the action involves very small hand movements. Such a big variance in these data samples increases the challenge for generative models to learn action state transition.

\subsection{Training Details for Visual Instruction Tuning and Action Frame Generation}
\label{sec:training_details}
\noindent\textbf{Visual Instruction Tuning.} We train the model with a batch size of 128 and a learning rate of $2\times10^{-5}$. Warm-up strategy and consine anneal~\cite{loshchilov2017sgdr} are also used in training. It takes 24 hours to train the model on 8 NVIDIA A100-SXM4-40GB for 3 epochs. AdamW~\cite{Loshchilov2017DecoupledWD} is adopted as the optimizer for training.

\noindent\textbf{Egocentric Action Frame Generation.} In training, we feed the input frame, ground truth frame (to obtain the gaussian noise through the diffusion process) together with the enriched action descriptions and VLLM embeddings into the model. We finetune the latent diffusion model with a batch size of 256 and an initial learning rate of $10^{-4}$ without warm-up strategy. Horizontal flipping is used as data augmentation. We train the model with optimizer AdamW~\cite{Loshchilov2017DecoupledWD} for 20,000 iterations on 8 NVIDIA A100-SXM4-40GB over 38 hours. In inference, we feed the input frame, a randomly-sampled gaussian noise as well as enriched action descriptions and VLLM embeddings into the latent diffusion model. We apply 100 denoising steps for each instance.

\subsection{Details about Classifier-free Guidance}
\label{sec:classifier_free_guidance}
We use classifier-free guidance for two conditions (following~\cite{brooks2023instructpix2pix}) by sharing the same guidance scale across the enriched action descriptions, VLLM image embedding and VLLM text embedding. As defined in \cref{sec:method_image_generation} in main paper, we use $\mathcal{C}$ to denote the three conditions and use $\mathcal{X}$ to denote the input frame. Specifically, we randomly set only the image conditioning $\mathcal{X}=\varnothing$ at a probability of 5\%, only the conditioning from VLLM $\mathcal{C}=\varnothing$ at a probability of 5\% and both $\mathcal{X}=\varnothing$ and $\mathcal{C}=\varnothing$ at a probability of 5\%. Then the score estimate of our model is formulated as
\begin{align}
    \Tilde{e}_{\theta}(z_t, \mathcal{X}, \mathcal{C}) &= e_\theta(z_t, \varnothing, \varnothing) \\
                                    &+ s_x\cdot(e_\theta(z_t, \mathcal{X}, \varnothing) - e_\theta(z_t, \varnothing, \varnothing)) \\
                                    &+ s_c\cdot(e_\theta(z_t, \mathcal{X}, \mathcal{C}) - e_\theta(z_t, \mathcal{X}, \varnothing)),
\end{align}
where $\theta$ refers to the parameters in the denoising UNet. $z_t$ is the noisy latent at timestep $t$, which is obtained by diffusion process in training and randomly initialized by a gaussian noise in inference. $s_x$ and $s_c$ are the guidance scales corresponding to the conditioning $\mathcal{X}$ and $\mathcal{C}$ respectively. In inference, we use $s_x=7.5$ and $s_c=1.5$ which are identical to the settings in InstructPix2Pix~\cite{brooks2023instructpix2pix}.

\begin{figure}[t]
\centering
\includegraphics[width=0.8\linewidth]{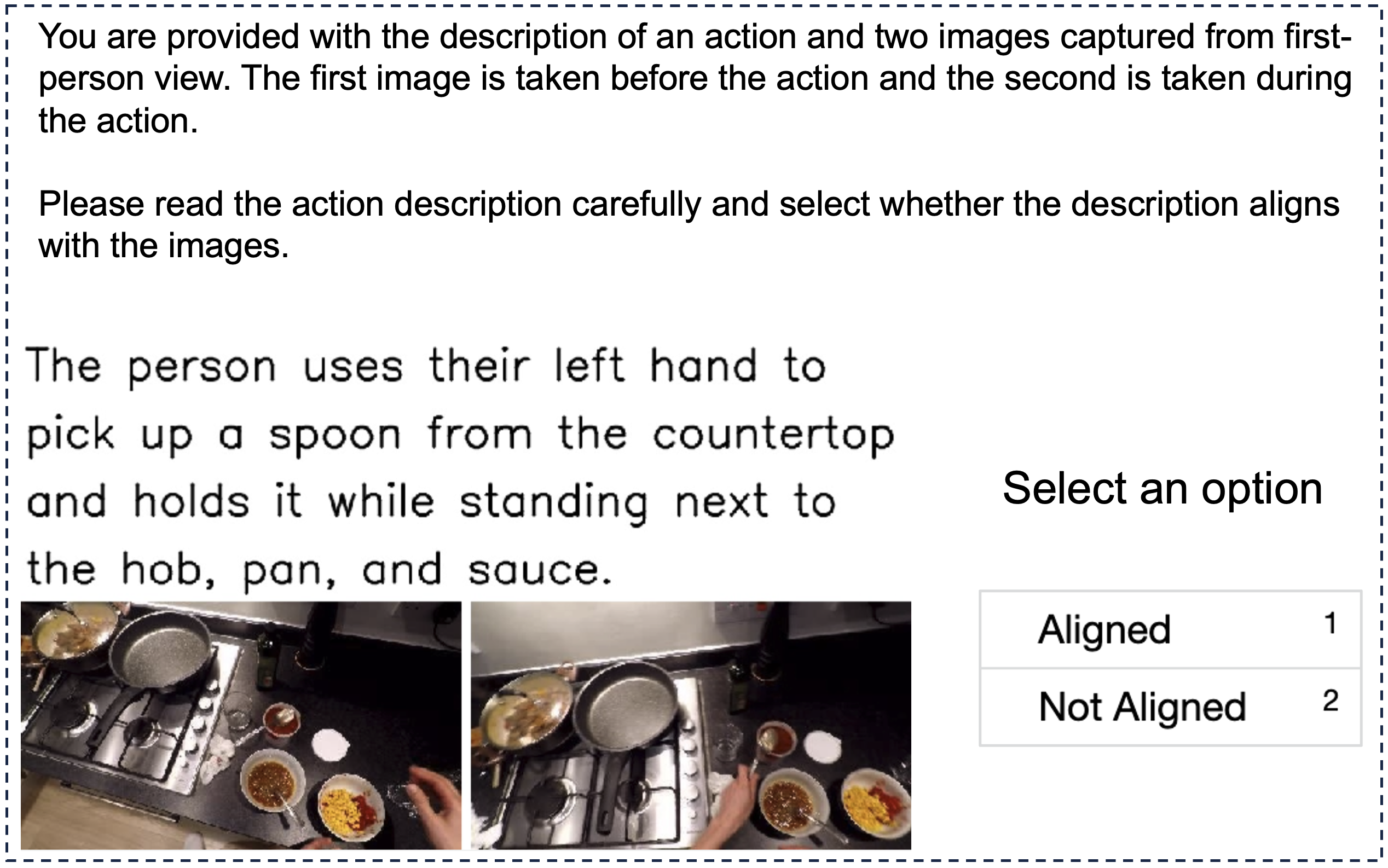}
\caption{The interface used for evaluation of enriched action descriptions. Both input and target frames are shown to the raters.}
\label{fig:interface_desc}
\end{figure}

\begin{figure}[t]
\centering
\includegraphics[width=0.8\linewidth]{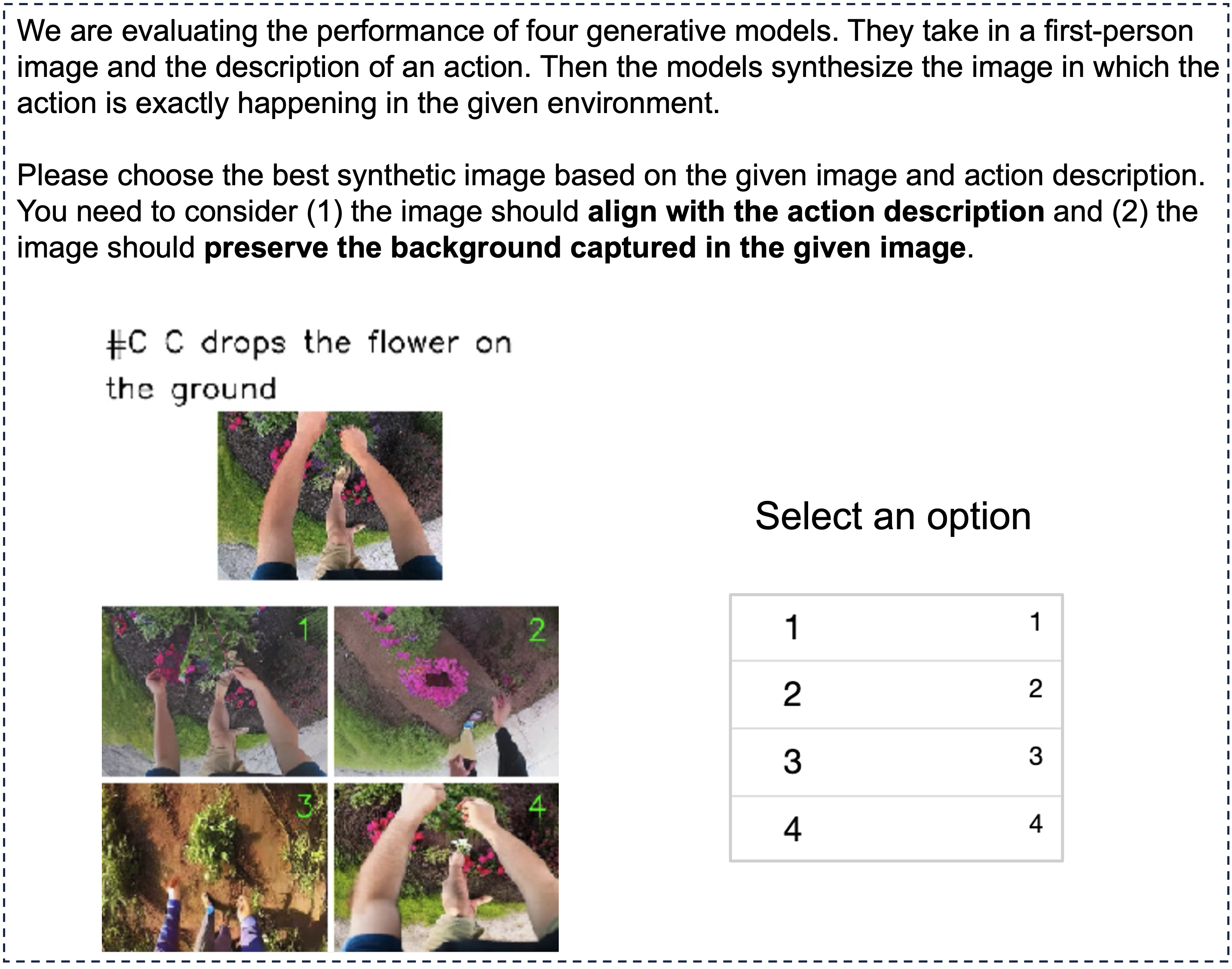}
\caption{The interface used for evaluation of generated action frames. The four generated frames are randomly shuffled to avoid potential bias.}
\label{fig:interface_gen}
\end{figure}

\subsection{Implementation Details of the Baseline Models}
\label{sec:baseline_models}
We compare our model with three prior image editing models -- ProxEdit~\cite{han2024proxedit}, SDEdit~\cite{meng2022sdedit} and InstructPix2Pix~\cite{brooks2023instructpix2pix}. We provide more implementation details for training the three models on our dataset. ProxEdit and SDEdit rely on a pre-trained diffusion model to edit the input image. Given the domain gap of the off-the-shelf diffusion models and our problem, we finetune a latent diffusion model with the egocentric action data using the default training hyper-parameters in~\cite{rombach2022high}. During finetuning, we use the action label as textual prompt and the target action frame as the ground truth. Then we implement ProxEdit and SDEdit on the finetuned diffusion model. In terms of InstructPix2Pix, we finetune it end to end with the egocentric action datasets using the same training hyper-parameters as our LEGO model (see \cref{sec:training_details}). Note that for all three baseline models, we use the short action labels as input rather than the detailed descriptions. The proposed action description enrichment is one of our key contributions, so we use it only for our model to show the benefit.

\subsection{Details and Interfaces for User Study}
\label{sec:user_study}
\noindent\textbf{User Study for the Enriched Action Descriptions.} In \cref{sec:exp_instruct_tuning} of the main paper, we apply the user study to evaluate the quality of enriched action descriptions from our instruction tuned VLLM and the off-the-shelf VLLM. We randomly sample 100 examples from the test set of each dataset. For each instance, we show the input frame, target frame and the action descriptions generated by VLLM. The rater is asked to select whether the description aligns with the two frames. We hire 5 raters for each instance on Amazon Mechanical Turk and finally get 500 samples for each dataset. The interface shown to raters is illustrated in \cref{fig:interface_desc}.

\noindent\textbf{User Study for Generated Action Frames.} The user interface for evaluation of generated action frames is illustrated in \cref{fig:interface_gen}. We show the input frame and shuffled outputs from the four models to raters. To make a fair comparison, we show the action label instead of the enriched action description because the baseline models only take original action labels as input.

\section{Limitation and Future Work}
\label{sec:future}
In this paper, we use an egocentric image to capture the user's environment contexts and generate the action frame to provide visual instructions. However, there are still some problems that are not explicitly solved by our method. We find it's hard for our model to associate the action with correct objects when there are too many irrelevant objects around. Synthesizing the diverse and complex hand-object interactions is also a big challenge especially when people are operating some machines. In addition, our work indicates a few valuable directions for the future study.

\begin{itemize}
    \item[$\bullet$] The embeddings from the visual large language model (VLLM) are fed into the UNet together with the CLIP based text representation as additional conditioning. How to leverage the VLLM embeddings more effectively in diffusion models deserves future study.
    \item[$\bullet$] Recognizing and localizing the objects that are relevant with the action descriptions in a chaotic environment may be a bottleneck for the application in real-world problems, which deserves more attention.
    \item[$\bullet$] It's difficult to synthesize correct interactions of hands and objects in some professional work, such as using a wood cutter, operating a lawn mower and sewing clothes on a sewing machine. How to combine affordance understanding with generative models may be a key step to address this problem.
    \item[$\bullet$] Existing automatic image-to-text similarity metric doesn't generalize well to the egocentric domain. We expect more investigation of better evaluation metrics for image-text alignment.
\end{itemize}

\section{Code and Data Release}
\label{sec:release}
We will release our code, pre-trained model weights, additional data annotations, train/test split, the enriched action descriptions and VLLM embeddings on the website (\url{https://bolinlai.github.io/Lego_EgoActGen/}) to the research community to facilitate future studies.

\end{document}